\documentclass[runningheads]{llncs}

\usepackage{eccv}

\usepackage{eccvabbrv}

\usepackage{amsthm}
\usepackage{graphicx}
\usepackage{booktabs}
\usepackage{multirow}
\usepackage{listings}
\usepackage{makecell}
\usepackage{algorithm}
\usepackage{wrapfig}
\usepackage{graphicx}
\usepackage{enumerate}
\usepackage{xcolor}
\usepackage{xspace}
\usepackage{algorithmic}
\usepackage{amssymb}
\usepackage{pifont}

\usepackage[accsupp]{axessibility}  
\usepackage{hyperref}

\usepackage{orcidlink}

\definecolor{GrayColor}{rgb}{0.5,0.5,0.5}
\definecolor{RedColor}{RGB}{255, 0, 84} 
\definecolor{BlueColor}{RGB}{5, 191, 219} 
\definecolor{GreenColor}{RGB}{22, 204, 0} 
\definecolor{YellowColor}{RGB}{225, 168, 0} 
\definecolor{PurpleColor}{RGB}{204, 51, 255} 

\newcommand{\textR}[1]{\textcolor{RedColor}{#1}}

\newcommand{\textG}[1]{\textcolor{GreenColor}{#1}}

\newcommand{\BGreen}[1]{\textcolor{GreenColor}{\bf #1}}

\newcommand{\STD}[1]{{\scriptsize$\pm${#1}}}

\renewcommand{\algorithmicrequire}{\textbf{Input:}}
\renewcommand{\algorithmicensure}{\textbf{Output:}}

\DeclareMathOperator*{\argmin}{arg\,min}

\begin{document}

\theoremstyle{definition}
\newtheorem{define}{Definition}

\title{Distill Gold from Massive Ores:
Bi-level Data Pruning towards Efficient Dataset Distillation} 

\titlerunning{Distill Gold from Massive Ores}


\author{Yue Xu\orcidlink{0000-0001-7489-7269}\inst{1} \and
Yong-Lu Li\orcidlink{0000-0003-0478-0692}\inst{1}\thanks{Corresponding author.} \and
Kaitong Cui\orcidlink{0009-0006-7709-8271}\inst{1} \and
Ziyu Wang\orcidlink{0009-0006-4480-6374}\inst{1} \and \\
Cewu Lu\orcidlink{0000-0003-1533-8576}\inst{1} \and
Yu-Wing Tai\orcidlink{0000-0002-3148-0380}\inst{2} \and
Chi-Keung Tang\orcidlink{0000-0001-7155-2919}\inst{3}
}

\authorrunning{Y.~Xu et al.}

\institute{Shanghai Jiao Tong University
\email{\{silicxuyue,yonglu\_li,ckt0704,wangxiaoyi2021,lucewu\}@sjtu.edu.cn, ckt0704@gmail.com} \\ \and
Dartmouth College \\
\email{yuwing@gmail.com} \\ \and
Hong Kong University of Science and Technology \\
\email{cktang@cs.ust.hk}
}

\maketitle

\begin{abstract}
    Data-efficient learning has garnered significant attention, especially given the current trend of large multi-modal models. Recently, dataset distillation has become an effective approach by synthesizing data samples that are essential for network training. However, it remains to be explored which samples are essential for the dataset distillation process itself.
    In this work, we study the data efficiency and selection for the dataset distillation task. By re-formulating the dynamics of distillation, we provide insight into the inherent redundancy in the real dataset, both theoretically and empirically. We propose to use the empirical loss value as a static data pruning criterion. 
    To further compensate for the variation of the data value in training, we find the most contributing samples based on their causal effects on the distillation.
    The proposed selection strategy can efficiently exploit the training dataset, outperform the previous SOTA distillation algorithms, and consistently enhance the distillation algorithms, even on much larger-scale and more heterogeneous datasets, \eg, full ImageNet-1K and Kinetics-400. 
    We believe this paradigm will open up new avenues in the dynamics of distillation and pave the way for efficient dataset distillation.
    \textbf{Our code is available on \url{https://github.com/silicx/GoldFromOres-BiLP}.}
\end{abstract}

\section{Introduction}

Data is the core of deep learning. In the era of large data and models~\cite{llama, laion, blip2} as their size and complexity continue to grow, {\em data-efficient} learning is increasingly crucial for achieving high performance with a limited computing budget.
Techniques such as pruning, quantization, and knowledge distillation have been developed to reduce model size without sacrificing performance. 
Recently, \textit{dataset distillation} \cite{DD} has emerged as a promising approach toward data-efficient AI, where a small and condensed dataset (namely {\em synthetic} dataset) is learned from the whole large dataset (namely {\em real} dataset), to maximize the performance of models trained on distilled synthetic data.

\begin{figure*}[t]
    \centering
    \subfloat{
        \includegraphics[width=0.55\textwidth]{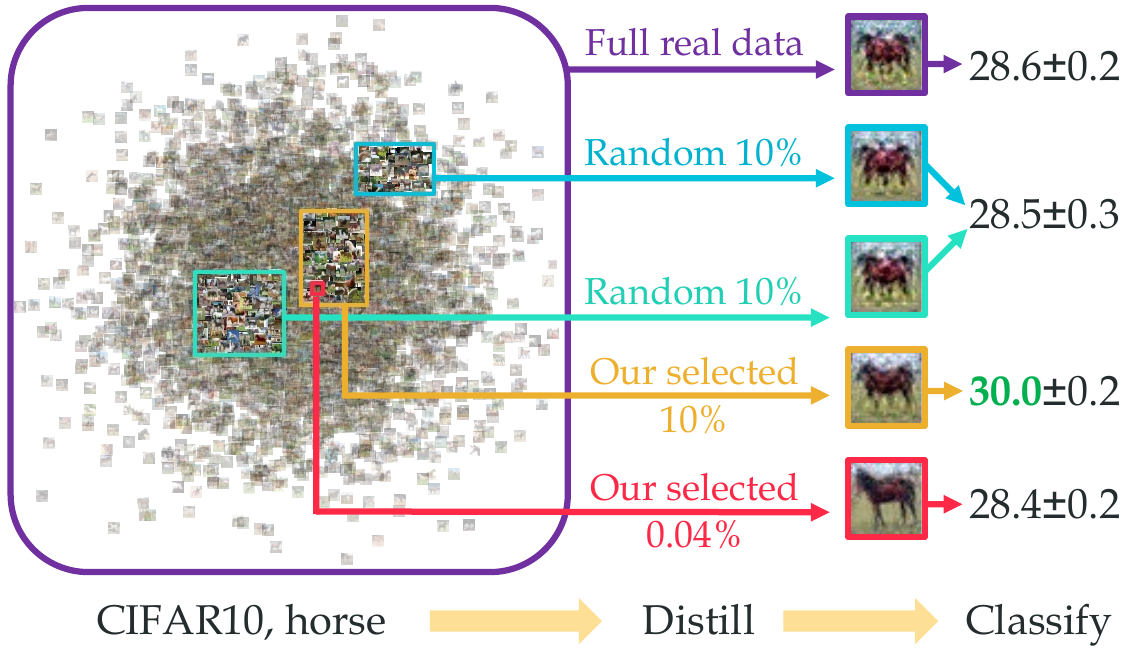}
    }
    \hspace{2ex}
    \subfloat{
	   \includegraphics[width=0.38\textwidth]{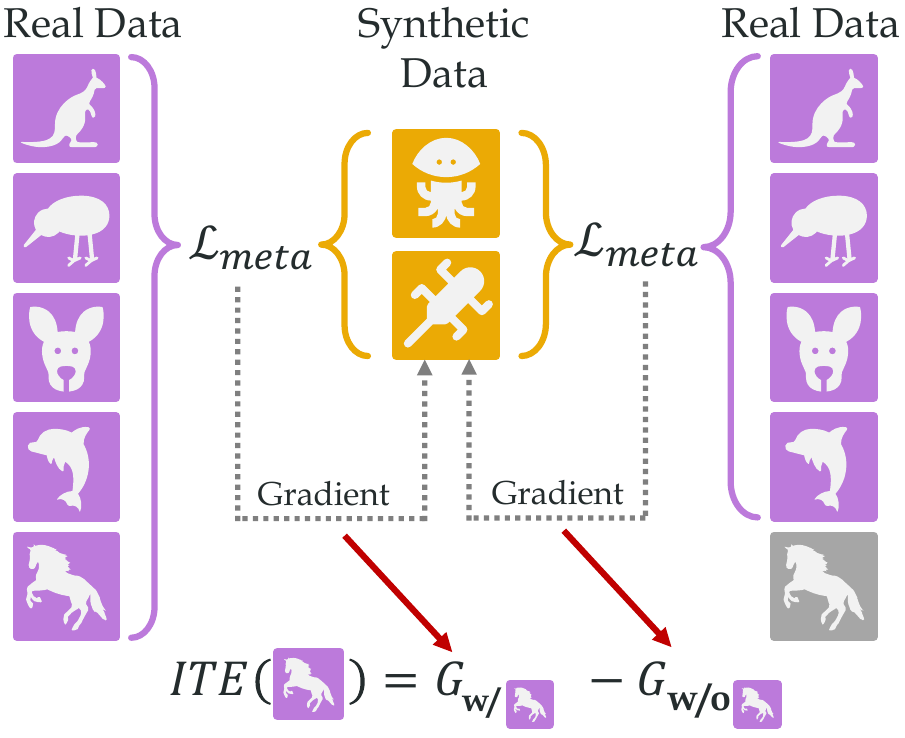}
    }
    \caption{(1) \textbf{Left}: \textit{Severe data redundancy} in the dataset distillation process. 
    Taking CIFAR10 and DC~\cite{DC} as an example, pruning 90\% of the real data does not reduce the performance.
    With our proposed empirical loss criterion, only 0.04\% real samples are sufficient for distillation and 10\% samples outperform the full real dataset. 
    (2) \textbf{Right}: Building upon the empirical loss, we propose to dynamically select samples by their ITE value, which is the difference of meta gradient with/without the sample.
    }
    \label{fig:teaser}
\end{figure*}

Currently, the existing distillation algorithms focus on the evolution of the matching strategies for real and synthetic data~\cite{DC,DM,MTT,shin2023loss,sajedi2023datadam} or refining and accelerating the bi-level optimization~\cite{DD,KIP,RFAD,RCIG}.
However, preliminary experiments show that the real dataset could be so redundant during the dataset distillation itself that pruning part of the dataset will not reduce the distillation performance (as shown in \cref{fig:teaser}).
Therefore, an appropriate approach to finding the most valuable real data would help the distillation algorithms to learn the most key data patterns and knowledge.
Currently, it is still under-explored how to analyze the samples' value on the dataset distillation task and efficiently select the real dataset.
Recently, YOCO~\cite{yoco} proposes an interesting task of pruning the synthetic dataset.
DREAM~\cite{DREAM} prunes the real data within each batch from the aspect of feature distribution. 
In comparison, we would principally analyze the intrinsic worth of each real data sample and its relevance to the distillation optimization process from a more holistic perspective.

In this work, we revisit the selection and utilization of real datasets for dataset distillation.
We first describe the dataset distillation as the matching of training dynamics on real and synthetic data, where the dynamics are formulated with the neural tangent kernel (NTK)~\cite{NTK}. Further analysis shows that the small scale of synthetic data motivates pruning those real data samples with large empirical loss values.
The stability of NTK for wide neural networks also allows for dropping these real samples at the beginning of dataset distillation, which indicates \textbf{redundancy} in real data and offers efficiency.
To support the theoretical analysis, we conduct comprehensive experiments of pruning real data before distillation. In most scenarios, pruning some real data does not reduce the distillation performance.
For instance, with CIFAR10 for DC~\cite{DC} and single instance-per-class (IPC=1), removing over \textbf{99.9}\% of the real data does not affect its performance.
We argue that leveraging redundancy in dataset distillation is more significant than regular machine learning applications since synthetic datasets usually possess low capacity, as shown in \cref{fig:teaser}.

However, pruning real data before distillation does not consistently enhance the distillation performance, which implies that the value or utility of the real samples could \textit{vary} during the training process.
To identify the importance of real samples, we measure their contribution to the learning of the synthetic data from the perspective of \textbf{causal effect}. We compare the synthetic gradient with or without each real sample, which is essentially the individual treatment effect (ITE) of the real samples for the dataset distillation task.
Based on the principles, we propose a new and concise real data sampling algorithm, which can be implemented as a plug-and-play for most dataset distillation methods.
In comprehensive experiments, our approach can thus efficiently use the key samples and consistently enhance the base algorithm.

Overall, our contributions are:
1) In-depth insight on data redundancy for dataset distillation.
2) Two core principles for real sample selection, supported both theoretically and empirically.
3) A plug-and-play selection strategy for practical dataset distillation to enhance SOTA methods.

\section{Related Work}

\textbf{Dataset Distillation} is a process of condensing a large dataset into a smaller and more representative dataset while maintaining the performance,
which has been applied to various domains including images, text~\cite{maekawa2023text,maekawa2024dilm}, videos~\cite{wang2023dancing}, graph~\cite{jin2022condensing,jin2022graph}, time series~\cite{liu2024time,ding2024time}, medical~\cite{li2020soft,li2022compressed} and multimodal data~\cite{wu2023multi,xu2024low}, etc.
Existing approaches can be roughly classified into: 
1) \textit{Meta-Model Matching} maintains the transferability of the synthetic data, by optimizing the loss on the original dataset of models trained on the synthetic data. 
In~\cite{DD} the task of data distillation and the meta-model matching framework was first proposed. 
In~\cite{KIP} kernel ridge regression was exploited to reduce computational cost and is further extended to infinite wide networks~\cite{KIP2}. In~\cite{FRePo} the optimization of synthetic data/classifier and feature extractor was separated. In~\cite{RCIG} the meta-gradient was computed by exploiting implicit gradients.
2) \textit{Gradient Matching} aligns the gradients of the synthetic and real datasets. It was proposed in~\cite{DC} and further improved in~\cite{DSA} to perform the same image augmentations on both the real and synthetic data.
3) \textit{Distribution Matching}, where~\cite{DM} matches feature distributions of the synthetic and real data, which is simple but effective. In~\cite{CAFE} layer-wise feature alignment and early exit conditions are used to promote it. In~\cite{IDM} it was further enhanced with regularizers and model pool.
4) \textit{Trajectory Matching:}
In~\cite{MTT} the authors proposed MTT to match the training trajectory of the model parameters and in~\cite{cui2022scaling} the memory consumption of MTT was reduced and label learning was used.
5) \textit{Factorization} of synthetic data can reduce the storage burden and share knowledge among instances. For example,
\cite{IDC} uses a strategy of putting multiple images on one synthetic sample. 
\cite{LinBa} decomposes the synthetic data to the linear network hallucinators and bases, while \cite{haba} uses a convolutional network. 
\cite{lee2022dataset} maintains a smaller base space to further reduce the storage.
\cite{frequency} employs frequency domain factorization.
6) \textit{Bayesian Pseudocoreset} is a family of algorithms that learn the synthetic data with Bayesian inference~\cite{manousakas2020bayesian, kim2022divergence, tiwary2023constructing}.
Beyond these categories, SRe2L~\cite{sre2l} utilizes a 3-stage learning paradigm to decouple the segregation of inner-loop and outer-loop optimization.

\textbf{Data Selection/Pruning} reduces the training data without significantly affecting performance. 
Classic data selection often calculates a scalar utility score for each sample based on predefined criteria~\cite{compactness, diversity, forgetfulness} and filters samples based on scores. 
Some data pruning methods also consider the interaction between samples. For example,
\cite{generalization} examines generalization influence to reduce training data, which aims to identify the smallest subset to satisfy the expected generalization ability.
In comparison, data distillation~\cite{DD} synthesizes new and smaller data, and significantly outperforms data pruning with the same data storage.

\section{Preliminaries}

\subsection{Formulation of Dataset Distillation}

Before delving deeper into the real data selection, we would investigate the training dynamics of dataset distillation. 
Given a real dataset $\mathcal{D}_r=\{x_r^{(i)},y_r^{(i)}\}_{i=1}^{M_r}$, dataset distillation is to learn a synthetic dataset $\mathcal{D}_s=\{x_s^{(i)},y_s^{(i)}\}_{i=1}^{{M_s}}$ that has smaller size ($M_s \ll M_r$) and could train a network to similar performance to full real dataset on specific task.
More formally, let $u=f(x;\boldsymbol{\theta})$ be a scalar output network~\footnote{
The analysis covers scalar functions, but can also generalize to vector functions.} and $\mathcal{L}(\mathcal{D},\boldsymbol{\theta})=\sum_{i=1}^M\ell(u^{(i)}, y^{(i)})$ be the loss function that measures the model empirical risk on a dataset $\mathcal{D}$. 
The dynamics of gradient descent optimization are described by differential equations: 
\begin{equation}
\begin{aligned}
& \dot{\boldsymbol{\theta}}_r
= - \frac{\partial \mathcal{L}(\mathcal{D}_r,\boldsymbol{\theta})}{\partial \boldsymbol{\theta}}
= - \sum_{i=1}^{M_r} \frac{\partial \ell(u_r^{(i)}, y_r^{(i)})}{\partial u_r^{(i)}}\frac{\partial u_r^{(i)}}{\partial \boldsymbol{\theta}},
\\
& \dot{\boldsymbol{\theta}}_s
= - \frac{\partial \mathcal{L}(\mathcal{D}_s,\boldsymbol{\theta})}{\partial \boldsymbol{\theta}}
= - \sum_{i=1}^{M_s} \frac{\partial \ell(u_s^{(i)}, y_s^{(i)})}{\partial u_s^{(i)}}\frac{\partial u_s^{(i)}}{\partial \boldsymbol{\theta}},
\end{aligned}
\end{equation}
where $\boldsymbol{\theta}_r(t)$ and $\boldsymbol{\theta}_s(t)$ are functions of timestamp $t$ which describe the training trajectory of parameters on the real or synthetic dataset. 
We omit ``$(t)$'' for clarity. 
$\dot{\boldsymbol{\theta}}$ indicates its derivative w.r.t $t$: $\frac{\partial\boldsymbol{\theta}}{\partial t}, \dot{\boldsymbol{\theta}}_s=\frac{\partial\boldsymbol{\theta}_s}{\partial t}$, and as a common process, the learning rate of gradient descent is absorbed to timestamp $t$ for simplicity.
The outputs of both networks on the real dataset are computed for evaluation:
\begin{equation}
\label{eq:network-formulation}
u_r^{(i)} = f(x_r^{(i)};\boldsymbol{\theta}_r), ~~u_s^{(i)} = f(x_s^{(i)};\boldsymbol{\theta}_s).
\end{equation}
Note that $u_r^{(i)}$ and $u_s^{(i)}$ are also function of $t$.

The current dataset distillation can be categorized into two types:
\begin{enumerate}[(1)]
    \item \textbf{Empirical Risk Minimization} includes DD~\cite{DD}, KIP~\cite{KIP,KIP2} and their variants~\cite{RCIG,RFAD}, which minimize the empirical risk on the real dataset of the network trained on synthetic data:
    \begin{equation}
        \mathcal{S} = \argmin_{\mathcal{S}} \sum_{i=1}^{M_r}\ell(u_s^{(i)}|_{t=+\infty}, y_r^{(i)}).
    \end{equation}
    \item \textbf{Training Dynamics Matching} is a simpler agent task for dataset distillation since the converged network $\boldsymbol{\theta}_s^{(i)}|_{t=+\infty}$ is usually intractable for bi-level optimization. Its ultimate target is to match the whole training trajectory on the real and synthetic datasets:
    \begin{equation}
    \label{eq:dynamics-match}
        \mathcal{S} = \argmin_{\mathcal{S}} D(u_s^{(i)}(t), u_r^{(i)}(t)), 
    \end{equation}
    where $D$ is a distance metric. The strategies include gradient trajectory matching~\cite{DC,IDC,DREAM,MTT}, feature matching~\cite{DM,CAFE,IDM}, loss matching~\cite{shin2023loss}, \textit{etc}.
\end{enumerate}
The generation-based methods~\cite{Generative,sre2l} are exceptions since they do not involve the network dynamics.

\subsection{Redundancy in Dataset Distillation}

Currently, the intrinsic value of each real sample under the dataset distillation scenario has not been thoroughly investigated, yet we observe the redundancy within the real dataset which underscores the potential significance of this research area.
We randomly drop some portions of the real data to find their maximum pruning ratio that could maintain the distillation accuracy.
Comprehensive experiments are conducted on various datasets, networks, distillation algorithms, initialization methods, and synthetic data sizes (represented by instance-per-class, IPC). 
We use the default parameters of each method, which are detailed in the supplementary material.
As shown in \cref{tab:rand-drop}, severe data redundancy widely exists in various dataset distillation settings. 
We will analyze and propose selection criteria dedicated to dataset distillation in the following.

\begin{table}[t]
    \centering
    \caption{Maximum pruning ratio with \textbf{random selection} among different datasets, algorithms and synthetic data size.}
    \resizebox{0.85\linewidth}{!}{
      \begin{tabular}{ll|ccccccc}
        \toprule
        Dataset  &IPC& DC~\cite{DC} & DSA~\cite{DSA} & DM~\cite{DM} & MTT~\cite{MTT} & IDC~\cite{IDC}  & IDM~\cite{IDM}\\
        \midrule[1.3pt] 
        \multirow{3}{*}{CIFAR10~\cite{cifar}}
                            & 1 & 90\% & 85\% & 85\% & 60\% & 50\% & 40\% \\
                            & 10& 70\% & 70\% & 60\% & 10\% & 30\% & 10\% \\
                            & 50& 50\% & 70\% & 50\% & 20\% & 40\% & 20\% \\
        \midrule
        \multirow{2}{*}{CIFAR100~\cite{cifar}}
                            & 1 & 50\% & 40\% & 70\% & 20\% & 60\% & 50\% \\
                            & 10& 50\% & 40\% & 50\% & 10\% & 50\% & 20\% \\
        \midrule
        \multirow{2}{*}{SVHN~\cite{svhn}}
                            & 1 & 60\% & 60\% & 85\% & 90\% & 60\% & 30\% \\
                            & 10& 80\% & 70\% & 95\% & 80\% & 30\% & 5\% \\
        \midrule
        \multirow{2}{*}{TinyImageNet~\cite{tinyimgnet}}
                            & 1 & 40\% & 30\% & 60\% & 50\% & 40\% & 60\%  \\
                            & 10& 50\% & 50\% & 50\% & 20\% & 30\% & 50\%  \\
        \bottomrule
      \end{tabular}
    }
    \label{tab:rand-drop}
\end{table}

\section{Methodology}

\subsection{Empirical Loss Pruning}

We study the real data selection of dataset distillation based on the previous formulation.
Considering that $\dot{u} = \langle \frac{\partial u}{\partial \boldsymbol{\theta}},\dot{\boldsymbol{\theta}} \rangle$, the dynamics of outputs can be depicted by differential equations:
\begin{equation}
\label{eq:dynamics-with-ntk}
\begin{aligned}
& \dot{u}_r^{(i)} =
- \sum_{j=1}^{M_r} \frac{\partial \ell(u_r^{(j)}, y_r^{(j)})}{\partial u_r^{(j)}}
\langle
\frac{\partial u_r^{(i)}}{\partial \boldsymbol{\theta}_r},
\frac{\partial u_r^{(j)}}{\partial \boldsymbol{\theta}_r}
\rangle
=
-\sum_{j=1}^{M_r} \frac{\partial \ell(u_r^{(j)}, y_r^{(j)})}{\partial u_r^{(j)}}\boldsymbol{K}_r^{(ij)},
\\
& \dot{u}_s^{(i)} =
- \sum_{j=1}^{M_s} \frac{\partial \ell(u_s^{(j)}, y_s^{(j)})}{\partial u_s^{(j)}}
\langle
\frac{\partial u_s^{(i)}}{\partial \boldsymbol{\theta}_s},
\frac{\partial u_s^{(j)}}{\partial \boldsymbol{\theta}_s}
\rangle
=
-\sum_{j=1}^{M_s} \frac{\partial \ell(u_s^{(j)}, y_s^{(j)})}{\partial u_s^{(j)}}\boldsymbol{K}_s^{(ij)},
\end{aligned}
\end{equation}
where $\boldsymbol{K}_r$ and $\boldsymbol{K}_s$ are neural tangent kernels~\cite{NTK} (NTK), and matching the derivatives $\dot{u}_r$ and $\dot{u}_s$ is a sufficient and necessary condition of target \cref{eq:dynamics-match}.

Based on the formulation of dataset distillation dynamics, we notice that $\boldsymbol{K}_r^{(ij)}$ is rank-deficient since synthetic data is much smaller than the real dataset. Since $\dot{u}^{i}$ is the linear combination of kernel matrix $\boldsymbol{K}$, the rank-deficiency of $\boldsymbol{K}_r^{(ij)}$ leads to the mismatch of complexity of $\dot{u}_s^{i}$ and $\dot{u}_r^{i}$, making synthetic dynamics $\dot{u}_s^{i}$ hard to approximate to $\dot{u}_r^{i}$.
So we argue that \textbf{the real dataset has significant redundancy for dataset distillation}.
To reduce the optimization difficulty for a more stable distillation process, and also for a more data-efficient distillation algorithm, the dataset distillation needs to select the most valuable real data samples. 
A more intuitive explanation is that the small synthetic data cannot memorize the larger scale real dataset, so selecting the important but easy samples will help the algorithm.

Furthermore, \cref{eq:dynamics-with-ntk} also reveals that samples with small gradient value $\|\frac{\partial \ell(u_r, y_r)}{\partial u_r}\|$ are more important for distillation, since pruning the samples with large gradient is equivalent to reduce the kernel rank of real data.
Note that for common loss functions like MSE or cross-entropy, the $\|\frac{\partial \ell(u_r, y_r)}{\partial u_r}\|$ has the same monotonicity to $\ell(u_r, y_r)$, so we directly \textbf{adopt the empirical loss $\ell(u_r, y_r)$ as our real data selection criterion} to avoid additional computation. Please refer to the supplementary for the proofs.

\subsubsection{Early Pruning.}
A principal property of the neural tangent kernels is they tend to be constant for wide neural networks. So in \cref{eq:dynamics-with-ntk}, the NTK for real data $\boldsymbol{K}_r$ is stable during training.
So the empirical loss can be regarded as a \textbf{static criterion} which does not vary during training.
In practice, to maximize the efficiency, we prune the real data in one time \textbf{before the beginning} of the distillation algorithm rather than select different samples at each iteration. 
The empirical loss of each sample is computed by training the network on the real dataset for multiple trials and taking the average loss value.

\subsubsection{Empirical Justification}
Besides the theoretical analysis, we conduct extensive experiments to support our selection strategy.
To examine the effectiveness of pruning, we compare the maximum pruning ratio that maintains the distillation performance, and a larger maximum pruning ratio indicates a larger redundancy.

\begin{figure}[t]
    \centering
    \includegraphics[width=0.9\linewidth]{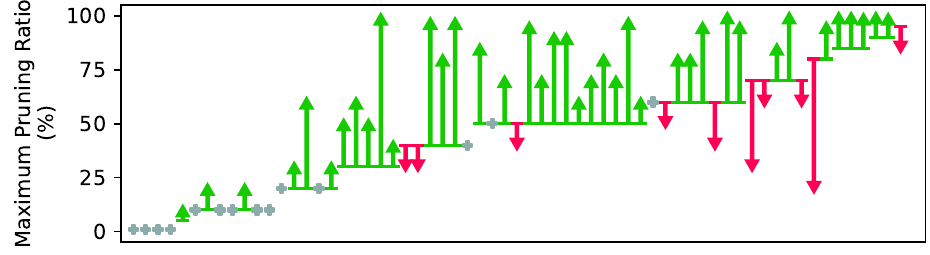}
    \caption{The comparison of maximum pruning ratio between random selection and empirical loss-based selection: the arrows point from random to the loss-based method. \textG{Green}: selection by loss is superior; \textR{Red}: selection by loss is inferior to random.}
    \label{fig:compare-loss-random-bar}
\end{figure}

Comprehensive comparison experiments on various datasets, networks,
distillation algorithms, initialization methods, and synthetic data sizes (represented by instance-per-class, IPC) are given in \cref{tab:compare-gamma-loss}. 
We take the mean and standard deviation of the accuracy of 5 random removal trials. We use the default parameters of each method, which are detailed in the supplementary.
The results show that severe data redundancy widely exists in various dataset distillation settings. 
For many datasets and algorithms, less than 30\% samples are sufficient for dataset distillation. 
We also visualize the comparison between random selection and empirical loss selection in \cref{fig:compare-loss-random-bar}. \textbf{Pruning with empirical loss consistently outperforms random selection}. 
We also find dropping data does not drop the cross-architecture transferability in supplementary.

\begin{table}[t]
    \centering
    \caption{Maximum pruning ratio with \textbf{empirical loss selection} among different datasets, algorithms and synthetic data sizes.} 
    \resizebox{0.9\linewidth}{!}{
      \begin{tabular}{l|l|ccccccc}
        \toprule
        Dataset &IPC& DC~\cite{DC} & DSA~\cite{DSA} & DM~\cite{DM} & MTT~\cite{MTT} & IDC~\cite{IDC} & RFAD~\cite{RFAD} & IDM~\cite{IDM} \\
        \midrule[1.3pt] 
        \multirow{3}{*}{CIFAR10~\cite{cifar}}
                            & 1 & 99.5\%& 99.5\%& 99.5\%& 60\% & 85\%  & 70\% & 30\%    \\
                            & 10&  30\% & 60\%  & 50\%  & 20\% & 50\%  & 20\% & 10\%    \\
                            & 50&  70\% & 85\%  & 40\%  & 20\% & 30\%  & 20\% & 30\%    \\
        \midrule
        \multirow{2}{*}{CIFAR100~\cite{cifar}}
                            & 1 & 95\%  & 97\%  &99.5\% & 60\% & 80\% &  80\% & 70\%    \\
                            & 10& 90\%  & 80\%  &  90\% & 20\% & 60\% &  80\% & 20\%    \\
        \midrule
        \multirow{2}{*}{SVHN~\cite{svhn}}
                            & 1 & 80\%  & 95\%  & 99\%  & 99\% & 40\% & 20\%  & 60\%    \\
                            & 10& 20\%  & 60\%  & 85\%  & 95\% & 50\% &  5\%  & 10\%    \\
        \midrule
        \multirow{2}{*}{TinyImageNet~\cite{tinyimgnet}}
                            & 1 & 97\%  & 99\%  & 99.5\%& 70\% & 40\% &    -  & 95\%    \\
                            & 10& 80\%  & 70\%  &   97\%& 30\% & 40\% &    -  & 60\%    \\
        \bottomrule
      \end{tabular}
    }
    \label{tab:compare-gamma-loss}
\end{table}

Surprisingly, we also observe that \textbf{dropping real data can sometimes improve performance}.
We show the best performance during data pruning in \cref{tab:loss-enhance-acc}. In almost all cases, the data selection can notably promote distillation accuracy. This implies that some samples may be ``detrimental'' to the distillation and the empirical loss is a proper criterion to cancel these negative impacts.
This observation inspires new approaches that leverage data utility and exploit all data samples in different quality, enabling future analysis of network dynamics and dataset distillation.

\begin{table*}[t]
    \centering
    \caption{Best performance with data dropping. The performance difference between the \textit{full real dataset} and \textit{pruned dataset} are shown in parentheses ($\dagger$: compare to our reproduced accuracy).}
    \resizebox{\textwidth}{!}{
      \begin{tabular}{l|c|cccccc}
        \toprule
        Dataset & IPC  &  DC~\cite{DC}  &  DSA~\cite{DSA}  &  DM~\cite{DM}  &  MTT~\cite{MTT} & RFAD~\cite{RFAD} & IDM~\cite{IDM}  \\
        \midrule[1.3pt]
\multirow{3}{*}{\makecell{CIFAR10\\ \cite{cifar}}}
            & 1  & 30.0\STD{0.2} ({\bf+1.7}) & 30.9\STD{0.1} ({\bf+2.6}) & 29.7\STD{0.3} ({\bf+3.7}) & 46.3\STD{0.8} ({+0.2})    & 53.7\STD{0.9} (+0.1)      & 46.3\STD{0.4} ({+\bf0.7})  \\
            & 10 & 44.9\STD{0.4} (+0.0)      & 52.4\STD{0.2} ({+0.2})    & 50.0\STD{0.2} ({\bf+1.1}) & 65.7\STD{0.3} ({\bf+0.4}) & 66.7\STD{0.2} ({+\bf0.4}) & 58.9\STD{0.3} ({+\bf0.3})  \\
            & 50 & 54.9\STD{0.5} ({\bf+1.0}) & 61.5\STD{0.7} ({\bf+0.9}) & 63.4\STD{0.2} ({\bf+0.4}) & 72.0\STD{0.4} ({\bf+0.4}) & 71.9\STD{0.2} ({+\bf0.8}) & 67.8\STD{0.2} ({+\bf0.3})  \\
        \midrule
        \multirow{2}{*}{\makecell{CIFAR100\\ \cite{cifar}}}
            & 1  & 14.1\STD{0.1} ({\bf+1.3}) & 15.6$\pm$0.1 ({\bf+1.7}) & 14.9$\pm$0.5 ({\bf+3.5}) & 24.6$\pm$0.4 ({\bf+0.3}) & 30.4$\pm$0.6 ({+\bf4.1}) & 25.9$\pm$0.3 ({+\bf5.8}) \\
            & 10 & 26.5\STD{0.3} ({\bf+1.3}) & 32.5$\pm$0.4 ({+0.2})    & 32.4$\pm$0.3 ({\bf+2.7}) & 40.1$\pm$0.5 (+0.0)      & 38.4$\pm$0.2 ({+\bf6.4}) & 45.9$\pm$0.1 ({+\bf0.8})  \\
        \midrule
        \multirow{2}{*}{SVHN~\cite{svhn}}
            & 1  & 32.2\STD{0.5} ({\bf+1.0}) & 28.9\STD{1.3} ({+0.1})    & 29.8\STD{0.5} ({\bf+6.3}$^\dagger$) & 43.0\STD{1.1} ({\bf+3.2}$^\dagger$) &       53.0\STD{0.2} ({\bf+0.8}$^\dagger$)& 71.9\STD{1.0} ({\bf+1.6}) \\
            & 10 & 76.2\STD{0.6} ({+0.1})    & 80.0\STD{0.8} ({\bf+0.8}) & 74.6\STD{0.3} ({\bf+0.9}$^\dagger$) & 78.1\STD{0.5} ({\bf+0.9}$^\dagger$) &       74.2\STD{0.2} ({+0.1}$^\dagger$)& 81.9\STD{0.5} ({+0.1})    \\
        \midrule
    \multirow{2}{*}{\makecell{TinyImage-\\Net~\cite{tinyimgnet}}}
        & 1  & 4.9\STD{0.1} ({\bf+0.2}$^\dagger$) & 4.3\STD{0.0} ({\bf+0.6}$^\dagger$) & 4.8\STD{0.1} ({\bf+0.9}) &  9.0\STD{0.4} ({\bf+0.2})  &        -     & 11.4\STD{0.2} ({\bf+1.3})  \\
        & 10 &12.8\STD{0.0} ({\bf+0.2}$^\dagger$) &14.8\STD{0.4} ({\bf+2.3}$^\dagger$) &17.5\STD{0.1} ({\bf+4.6}) & 23.8\STD{0.3} ({\bf+0.6})  &        -     & 22.9\STD{0.4} ({\bf+1.0})  \\
        \bottomrule
      \end{tabular}
    }
    \label{tab:loss-enhance-acc}
\end{table*}

\subsection{Causal Effect on Synthetic Data}
\label{subsec:syn-prune}

However, the pruning lacks availability in some scenarios when IPC is large on certain algorithms, indicating that sample importance may vary at different stages of the training process; some initially pruned samples can be beneficial in the latter stage.
So to take a step further, we investigate the real samples' contribution to the synthetic data during training as compensation for the static empirical loss criterion.

We first generalize the optimization targets of various dataset distillations into a meta-loss function $\mathcal{L}_\text{meta}(\mathcal{D}_r, \mathcal{D}_s)$.
\Eg, the meta-loss of DC~\cite{DC} is the cosine distance between the gradients of real and synthetic; the meta-loss of DM~\cite{DM} is the MMD between real and synthetic features.
The synthetic data $\mathcal{D}_s$ is learned by gradient descent on the meta-loss function. 
To examine the causal effects of real samples on the updating and learning of synthetic data, we could observe the consequence on the meta gradient after applying \textbf{causal intervention} on the real dataset, \ie, removing a certain real sample.
More specifically, we regard the presence of real samples $x_r\in \mathcal{D}_r$ as some binary ``treatment'' ($T$), and the meta gradient of synthetic data $ \frac{\partial L_\text{meta}}{\partial \mathcal{D}_s}$  as some ``effect'' ($Y$).
Thus for a certain individual $\mathcal{D}_s$, the causal effect of each real sample to the distillation can be quantized by the individual treatment effect (ITE)~\cite{rubin2005causal}:
\begin{equation}
\label{eq:ite-original}
    ITE(x_r)=Y_{T=1}-Y_{T=0}=\frac{\partial L_{meta}(\mathcal{D}_r,\mathcal{D}_s)}{\partial \mathcal{D}_s}-\frac{\partial L_{meta}(\mathcal{D}_r \setminus \{x_r\},\mathcal{D}_s)}{\partial \mathcal{D}_s},
\end{equation}
which is the difference between the gradient on the synthetic dataset with or without the real sample $x_r$, and can be obtained on any dataset distillation algorithm by computing two meta gradients for each real sample.
We use the L2-norm of ITE gradient $\|ITE(x_r)\|_2$ as the causal criterion of a real sample $x_r$.
We remove the samples with both the \textit{smallest} and the \textit{largest} ITE criterion values since the small ITE value samples contribute little to the synthetic learning, while the large ITE value samples could be outliers for the distillation, which enhances the variance and instability during training.
The ablation analysis in \cref{sec:prune-component} also shows the samples with intermediate-level ITE are more important to the distillation task.

Due to the heavy computation of the meta gradient, we apply three optimization techniques for applicable efficiency.

\noindent\textbf{Taylor Approximation of ITE}.
The causal criterion needs additional meta gradient computations for each of the $K$ real samples (the second term in \cref{eq:ite-original}), therefore increasing the training time by nearly $K$ times. 
Thankfully, we can reduce the computation burden on a special family of meta loss functions that are \textit{additive variable separable}: $\mathcal{L}_{meta}(G, \mathcal{D}_s),~G=\sum_{x\in \mathcal{D}_r}g(x)$, 
which covers most of the meta matching algorithms, \eg, $g(x)$ indicates per-sample gradient for DC~\cite{DC}, sample feature vector for DM~\cite{DM}.
In this case, we could leverage the Taylor approximation for ITE value:

\begin{equation}
\label{eq:taylor-approx}
\begin{aligned}
    ITE(x_r) &= 
    \frac{\partial}{\partial \mathcal{D}_s} \left[ \mathcal{L}_{meta}(G, \mathcal{D}_s) - \mathcal{L}_{meta}(G-g(x_r), \mathcal{D}_s) \right]
    \\
    &= \frac{\partial}{\partial \mathcal{D}_s} \left[ \left(\frac{\partial L_{meta}(G, \mathcal{D}_s)}{\partial G}\right)^\top  g(x_r)+ h.o.t \right] 
    \\ 
    &\approx \frac{\partial^2 L_{meta}(G, \mathcal{D}_s)}{\partial \mathcal{D}_s ~\partial G} \cdot g(x_r),
\end{aligned}
\end{equation}
where we vectorize $G$, $g(x)$, and $\mathcal{D}_s$ for clarity.
The first term is part of the Hessian matrix of $\mathcal{L}_{meta}$ and not related to specific real sample $x_r$ so it could be computed in one backward pass.
The second term has to be computed during the forward pass of meta loss.
So we could compute the ITE values for the real batch with one meta gradient computation and a matrix multiplication, which significantly reduces the pruning time.
We also analyze the error of the approximation in the supplementary and ignoring high order terms only brings $3\%$ shifting of the sample ranking that is negligible for pruning.

\noindent\textbf{Estimation of Global ITE Distribution}.
We hope to prune the samples with small ITE values among the full real dataset, but it is computationally expensive to exhaustively evaluate ITE values of the \textit{full} dataset at each iteration, especially with mini-batch optimization methods.
Inspired by batch normalization~\cite{batchnorm}, we maintain the global running mean $\hat\mu$ and variance $\hat\sigma^2$ of ITE values of the real dataset. For each mini-batch, we compute the mean $\mu_B$ and variance $\sigma^2_B$ within the batch, and update the global statistics by:
\begin{equation}
\label{eq:running-stats}
    \hat\mu\leftarrow (1-\eta)\hat\mu + \eta\mu_B,\quad  \hat\sigma^2\leftarrow (1-\eta)\hat\sigma^2 + \eta\sigma^2_B.
\end{equation}
Then, given a pruning ratio $\beta$, we find the two quantile points $z_\frac{\beta}2$ and $z_{1-\frac{\beta}2}$ of the normal distribution $z\sim\mathcal{N}(\hat\mu, \hat\sigma^2)$, and select the samples with ITE value within $(z_\frac{\beta}2, z_{1-\frac{\beta}2})$, so that the overall selection rate is $1-\beta$.

\noindent\textbf{Lazy Selection}.
We prune and update the real data batch every several iterations to reduce the overhead of selection.

The three optimizations could reduce the ITE computation to 1/60,000 of the original time. Overall, our method only increases the training time by $7\%$.

\subsection{Bi-level Data Pruning for Dataset Distillation}
\label{subsec:prune-alg}

In the analysis and discussion above, we propose two levels of data selection: the \textit{preemptive} pruning by sample-wise empirical loss which is applied \textit{before} the distillation process, and \textit{adaptive} pruning by causal effect metric which is used \textit{during} the distillation.
We combine the two levels to propose a plug-and-play data pruning algorithm whose pseudo-code is shown in \cref{alg:bilp}.

\begin{algorithm}[t]
    \renewcommand{\algorithmicrequire}{\textbf{Input:}}
    \renewcommand{\algorithmicensure}{\textbf{Output:}}
    \caption{\textbf{Bi}-\textbf{L}evel Data \textbf{P}runing (\textbf{BiLP})}
    \label{alg:bilp}
    \begin{algorithmic}[1]
    \REQUIRE Real dataset $\mathcal{D}_r$, preemptive pruning rate $\alpha$, adaptive pruning rate $\beta$, meta loss function $\mathcal{L}_{meta}$
    \ENSURE Synthetic dataset $\mathcal{D}_s$,
    \STATE Train networks on full dataset $\mathcal{D}_r$ and record the average per-sample empirical loss $\ell(x_r),~\forall x_r \in \mathcal{D}_r$ with \cref{eq:network-formulation},
    \STATE $\mathcal{D}'_r \leftarrow \{x_r ~|~ \ell(x_r)<\tau, x_r \in \mathcal{D}_r\},~s.t.~~|\mathcal{D}'_r|=(1-\alpha)|\mathcal{D}_r|$ (\textbf{preemptive pruning}),
    \STATE Initialize synthetic dataset $\mathcal{D}_s$,
    \STATE Initialize running mean $\hat\mu$ and variance $\hat\sigma^2$ of ITE,
    \REPEAT
        \STATE Sample a mini-batch $\mathcal{B}_r$ from real dataset,
        \STATE Compute meta loss $\mathcal{L}_{meta}(\mathcal{B}_r, \mathcal{D}_s)$,
        \STATE Compute meta loss after causal intervention $\mathcal{L}_{meta}(\mathcal{B}_r\setminus\{x_r\}, \mathcal{D}_s),~\forall x_r\in \mathcal{B}_r$,
        \STATE Compute ITE values of real data $ITE(x_r),~\forall x_r\in \mathcal{B}_r$,
        \STATE Update running stats $\hat\mu$, $\hat\sigma^2$ with \cref{eq:running-stats},
        \STATE $\mathcal{B}'_r\leftarrow \{x_r ~|~ z_{\frac\beta2}<ITE(x_r)<z_{1-\frac\beta2},z\sim\mathcal{N}(\hat\mu, \hat\sigma^2)\}$ (\textbf{adaptive pruning}),
        \STATE Update synthetic data: $\mathcal{D}_s \leftarrow \mathcal{D}_s-\frac{\partial}{\partial \mathcal{D}_s}\mathcal{L}_{meta}(\mathcal{B}'_r, \mathcal{D}_s)$.
    \UNTIL{convergence}
    \end{algorithmic}
\end{algorithm}

\section{Experiments}

\subsection{Datasets and Metrics}
In this work, the experiments are conducted on common datasets for dataset distillation task, including CIFAR10~\cite{cifar} (60,000 32x32 images in 10 classes), CIFAR100~\cite{cifar} (60,000 32x32 images in 100 classes), SVHN~\cite{svhn} (over 99,000 32×32 images in 10 classes) and TinyImageNet~\cite{tinyimgnet} (100,000 64x64 images in 200 classes).
We report the top-1 classification accuracy.

\subsection{Implementation Details}

We apply BiLP on gradient-matching based SOTAs DC~\cite{DC} and IDC~\cite{IDC}, so $g(x)$ in \cref{eq:taylor-approx} indicates the per-sample gradient of the classification task.
The multi-formation factor is 2 by default to fairly compare to the IDC baseline, and we also show the results with factor 3 (BiLP+IDC(x3) in \cref{tab:main-result}). 
All the experiments including the efficiency analysis are conducted on one single RTX4090. For more details please refer to the supplementary.

\begin{table*}[t]
\centering
\caption{Dataset distillation performance of state-of-the-art and the proposed BiLP.} 
\resizebox{\textwidth}{!}{
  \begin{tabular}{l|ccc|cc|ccc}
    \toprule
    Method & \multicolumn{3}{c|}{CIFAR10~\cite{cifar}} & \multicolumn{2}{c|}{CIFAR100~\cite{cifar}} & \multicolumn{3}{c}{SVHN~\cite{svhn}} \\
    IPC           &   1         &   10        &   50       &   1       &       10   &       1   &       10   &        50  \\
    \midrule[1.3pt]
    Full Dataset  &  \multicolumn{3}{c|}{84.8$\pm$0.1} & \multicolumn{2}{c|}{56.2$\pm$0.3} & \multicolumn{3}{c}{95.4$\pm$0.1} \\
    \midrule
    Random        & 14.4\STD{2.0} & 26.0\STD{1.2} & 43.4\STD{1.0} & 4.2\STD{0.3} & 14.6\STD{0.5} & 14.6\STD{1.6} & 35.1\STD{4.1} & 70.9\STD{0.9}  \\
    Herding       & 21.5\STD{1.2} & 31.6\STD{0.7} & 40.4\STD{0.6} & 8.4\STD{0.3} & 17.3\STD{0.3} & 20.9\STD{1.3} & 50.5\STD{3.3} & 72.6\STD{0.8}  \\
    \midrule
    DC~\cite{DC}       & 28.3\STD{0.5} & 44.9\STD{0.5} & 53.9\STD{0.5} & 12.8\STD{0.3} & 25.2\STD{0.3} & 31.2\STD{1.4} & 76.1\STD{0.6} & 82.3\STD{0.3}   \\
    KIP~\cite{KIP}     & 49.9\STD{0.2} & 62.7\STD{0.3} & 68.6\STD{0.2} & 15.7\STD{0.2} & 28.3\STD{0.1} & 57.3\STD{0.1} & 75.0\STD{0.1} & 80.5\STD{0.1}   \\
    MTT~\cite{MTT}     & 46.3\STD{0.8} & 65.6\STD{0.7} & 71.6\STD{0.2} & 24.3\STD{0.3} & 40.1\STD{0.4} &      -       &       -      &       -        \\
    FRePo~\cite{FRePo} & 46.8\STD{0.7} & 65.5\STD{0.4} & 71.7\STD{0.2} & 28.7\STD{0.1} & 42.5\STD{0.2} &      -       &       -      &       -        \\
    HaBa~\cite{haba}   & 48.3\STD{0.8} & 69.9\STD{0.4} & 74.0\STD{0.2} & 33.4\STD{0.4} & 40.2\STD{0.2} & 69.8\STD{1.3} & 83.2\STD{0.4} & 88.3\STD{0.1}  \\
    IDC~\cite{IDC}     & 50.6\STD{0.4} & 67.5\STD{0.5} & 74.5\STD{0.1} &       -       & 45.1\STD{0.4} & 68.5\STD{0.9} & 87.5\STD{0.3} & 90.1\STD{0.1}  \\
    RFAD-NN~\cite{RFAD}& 53.6\STD{1.2} & 66.3\STD{0.5} & 71.1\STD{0.4} & 26.3\STD{1.1} & 33.0\STD{0.3} & 52.2\STD{2.2} & 74.9\STD{0.4} & 80.9\STD{0.3}  \\
    IDM~\cite{IDM}     & 45.6\STD{0.7} & 58.6\STD{0.1} & 67.5\STD{0.1} & 20.1\STD{0.3} & 45.1\STD{0.1} &    -    &       -      &       -       \\
    Zhang~\etal~\cite{zhang2023accelerating} & 49.2  & 67.1 & 73.8     & 29.8         & 45.6         &    -    &       -      &       -        \\
    DREAM~\cite{DREAM} & 51.1\STD{0.3} & 69.4\STD{0.4} & 74.8\STD{0.1} & 29.5\STD{0.3} & 46.8\STD{0.7} & 69.8\STD{0.8} & 87.9\STD{0.4} & 90.5\STD{0.1} \\
    PDD~\cite{vodka}  &    -        & 67.9$\pm$0.2 & 76.5$\pm$0.4 &       -      & 45.8$\pm$0.5 &    -    &       -      &       -        \\
    \midrule
    BiLP+DC      & 30.5\STD{0.3} & 45.2\STD{0.4} & 54.9\STD{0.3} & 13.7\STD{0.7} & 26.0\STD{0.5} & 32.2\STD{0.3} & 76.4\STD{0.5} & 82.8\STD{0.6}  \\
    BiLP+IDC     & 51.5\STD{0.3} & 69.4\STD{0.5} & 75.4\STD{0.2} & 30.1\STD{0.4} & 47.2\STD{0.6} & 70.3\STD{0.6} & 88.3\STD{0.1} & 90.8\STD{0.4}  \\
    BiLP+IDC (x3)& {\bf55.9\STD{0.5}} & {\bf69.8\STD{1.1}}  & {\bf76.9\STD{0.9}} & {\bf34.0\STD{0.7}} & {\bf48.0\STD{1.0}} & {\bf77.2\STD{0.6}} & {\bf88.7\STD{0.4}} &  {\bf91.0\STD{0.7}}    \\ 
    \bottomrule
  \end{tabular}
}
\label{tab:main-result}
\end{table*}

\subsection{Results}

We compare our BiLP to various baselines in \cref{tab:main-result}. 
Our selection strategy could notably enhance the current distillation algorithms. On average, BiLP consistently enhances DC and IDC by $0.8\%$ and $1.2\%$ (BiLP+DC and BiLP+IDC), especially on more diversified dataset CIFAR100.
Moreover, with a larger multi-formation factor (BiLP+IDC x3), our method could surpass most of the state-of-the-art. 
Note that our method is also efficient due to the preemptive pruning, \eg, BiLP on CIFAR100 only required 50\% samples of the real dataset.
This experiment shows the feasibility of embedding the data selection mechanism into the current distillation paradigm to boost performance and enhance efficiency, especially the preemptive pruning before the distillation.

\subsection{Ablation Study}

\subsubsection{Pruning Criteria}
\label{sec:prune-component}
We analyze the impact of different pruning criteria in Table~\ref{tab:abl-prune}, involving the preemptive pruning by empirical loss and adaptive pruning with causal effects in two directions (prune the samples with large ITE or small ITE).
The results indicate that across all datasets and IPC settings, the full BiLP consistently achieves the highest accuracy.
In addition, both preemptive and adaptive pruning brings performance gain.
Interestingly, pruning large ITE values demonstrates a slightly better performance than pruning small ITE values, achieving the highest accuracy of 69.2\%. This could imply that removing larger ITE may be more beneficial in scenarios where the synthetic data's complexity is limited.
But overall, applying pruning of both large and small ITE would outperform any pruning in a single direction.

\begin{table*}[t]
\centering
\caption{Ablation study on the pruning criteria on CIFAR10 and CIFAR100.}
\resizebox{0.7\textwidth}{!}{
  \begin{tabular}{l|ccccc}
    \toprule
    Preemptive Pruning &  \ding{55}    &  \checkmark   &   \checkmark  &  \checkmark   &       \checkmark    \\
    Prune Small ITE    &  \ding{55}    &  \ding{55}    &   \ding{55}   &  \checkmark   &       \checkmark    \\
    Prune Large ITE    &  \ding{55}    &  \ding{55}    &   \checkmark  &   \ding{55}   &       \checkmark    \\
    \midrule
    CIFAR10, IPC=1     & 50.6\STD{0.4} & 51.0\STD{0.3} & 51.2\STD{0.3} & 51.0\STD{0.3} & {\bf 51.3\STD{0.3}}  \\
    CIFAR10, IPC=10    & 67.5\STD{0.5} & 68.5\STD{0.1} & 68.9\STD{0.5} & 68.7\STD{0.4} & {\bf 69.2\STD{0.1}}  \\
    CIFAR100, IPC=1    & 28.2\STD{0.6} & 29.1\STD{0.8} & 29.8\STD{0.2} & 29.4\STD{0.2} & {\bf 30.1\STD{0.4}}  \\
    \bottomrule
  \end{tabular}
}
\label{tab:abl-prune}
\end{table*}

\begin{table*}[t]
\centering
\caption{Comparison of training time with ITE computation methods. We report the training seconds per iteration (s/iter) on CIFAR10 and IPC=10.}
\resizebox{0.68\textwidth}{!}{
  \begin{tabular}{l|cccc}
    \toprule
    Running Stats Estimation&  \ding{55}   &  \checkmark   &   \checkmark  &  \checkmark   \\
    Taylor Approximation    &  \ding{55}   &   \ding{55}   &   \checkmark  &  \checkmark   \\
    Lazy Selection          &  \ding{55}   &   \ding{55}   &   \ding{55}   &  \checkmark   \\
    \midrule
    ITE Computation (s/iter)&   ~1799.22~  &    ~23.03~    &     ~0.27~    &     ~\textbf{0.03}~      \\
    Total Iteration (s/iter)&   ~1799.53~  &    ~23.38~    &     ~0.65~    &     ~\textbf{0.47}~      \\
    \bottomrule
  \end{tabular}
}
\label{tab:abl-ite-time}
\end{table*}

\subsubsection{Time Complexity}
Since the computation time of ITE would pose a major bottleneck to the algorithm efficiency of BiLP, we offer a comparative analysis of the training time per iteration for different optimization levels in Table \ref{tab:abl-ite-time}. The experiment is conducted on the CIFAR10 dataset with IPC=10 on IDC~\cite{IDC}.
Without any optimization, the computation of ITE takes 5,800x more time than the rest steps of the distillation due to its computational intensity on the multiple meta gradients.
The running estimation of global ITE distribution would help the mini-batch-based optimization.
The Taylor approximation could effectively reduce the computation time by 98.8\%.
With the lazy selection (update the data per 10 iterations like~\cite{DREAM}), our full optimization techniques could reduce the training overhead to ignorable 7\% of the original algorithm.

\begin{figure}[t]
    \centering
    \includegraphics[width=\linewidth]{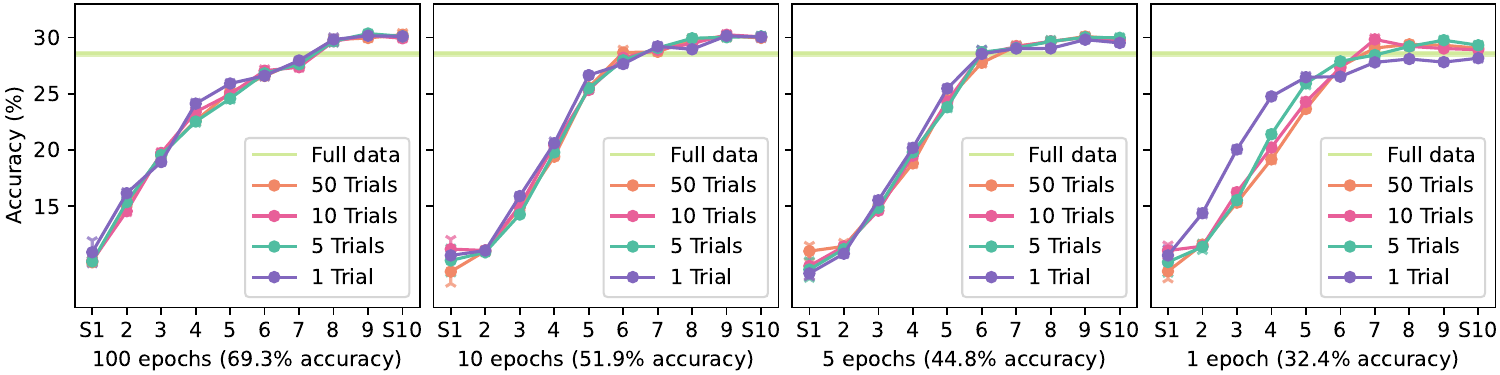}
    \caption{Stratified distillation performance of different classification trials and epochs numbers. 
    An ideal stratification yields a monotonously increasing performance curve.}
    \label{fig:loss-epoch-and-repeats}
\end{figure}

\subsubsection{Computation of Empirical Loss Criterion}
\label{sec:early-loss-ablation}
For the preemptive pruning, we train the networks on the full dataset and take the per-sample loss after convergence as the empirical loss criterion.
However, we find that the loss values in very early classification epochs are also informative enough for data selection, probably as the early dynamics of samples can reflect their training difficulty and importance. Moreover, the number of trials to obtain loss values has little influence on the selection performance.

We use a stratified experiment to examine the two factors: we sort the samples from large to small by the loss value and stratify them into 10 partitions S1$\sim$S10, where S1 contains the samples with the largest loss values.
We take the average loss curves of 1, 5, 10, and 50 trials of classification, and take the mean loss value for the first 1, 5, 10, and 100 epochs when the training converges at around 100 epochs.
As shown in \cref{fig:loss-epoch-and-repeats} by the empirical results with DC~\cite{DC} on CIFAR10, IPC=1, most loss values produce a good stratification and could distinguish the important samples unless only train the network for 1 epoch.
Therefore we argue that \textbf{early epoch loss in very few trials is also accurate for selection}, which can be utilized with reduced computational burden of generating loss values, and can also be ignored or naturally embedded in the distillation process itself, extending our paradigm to broader applications.

\subsection{Extended Discussion and Limitation}

\subsubsection{Efficient Distillation of Large-scale Datasets}

Our data utility paradigm can be efficiently extended to larger-scale and more heterogeneous datasets.
We apply the data utility selection to the distillation of ImageNet-1K~\cite{imagenet} and scale up to IPC=50, and also the large-scale video dataset Kinetics-400~\cite{kinetics} (detailed in supplementary). The results are listed in \cref{tab:large-data}.
Most methods struggle with high IPC due to demanding GRAM, except DM which allows class-separate training. MTT is extremely expensive for large-scale data due to its expert training.
Our BiLP could mitigate training costs by its preemptive pruning, which significantly reduces the training time by at most 60\% while maintaining or enhancing the performance.
It is especially suitable for large-scale scenarios when the size increases, whose signal-to-noise ratio continues to decrease.

\begin{table*}[t]
    \centering
    \caption{Dataset distillation on large-scale image and video datasets (\textit{est.}: estimated).}
    \resizebox{0.85\textwidth}{!}{
      \begin{tabular}{l|c|l|cc|cc|c}
        \toprule
        \multirow{2}{*}{Dataset} & \multirow{2}{*}{IPC} & \multirow{2}{*}{Algorithm}
                & \multicolumn{2}{c|}{Full Data} & \multicolumn{2}{c|}{BiLP} & \multirow{2}{*}{Random Real}  \\
        &   &   & Accuracy & Training Time & Accuracy & Training Time & \\
        \midrule[1.3pt]
        \multirow{5}{*}{\makecell{ImageNet-1K~\cite{imagenet} \\(1.3M images)}} 
        &  \multirow{3}{*}{1}  & DC~\cite{DC}   & 1.79$\pm$0.04  &    23.6h               & \textbf{2.02$\pm$0.10} & \textbf{17.3h} &  \multirow{3}{*}{0.43$\pm$0.02} \\
        &                      & DM~\cite{DM}   & 1.58$\pm$0.11  &    22.4h               & \textbf{1.95$\pm$0.12} & \textbf{9.9h}  &        \\
        &                      & MTT~\cite{MTT} &  -             & 205.0h (\textit{est.}) & \textbf{2.10$\pm$0.08} & \textbf{31.2h} &        \\   
        \cmidrule(l){2-8}
        &         10           & DM~\cite{DM}   & 3.86$\pm$0.16  & 24.7h & \textbf{5.21$\pm$0.11} & \textbf{15.3h}  & 1.57$\pm$0.21  \\
        \cmidrule(l){2-8}
        &         50           & DM~\cite{DM}   & 8.22$\pm$0.86  & 35.0h & \textbf{9.41$\pm$0.38} & \textbf{20.6h}  & 5.29$\pm$0.70  \\         
        \midrule
        \multirow{3}{*}{\makecell{Kinetics-400~\cite{kinetics} \\ (300K videos)}}
        & \multirow{2}{*}{1} & DM~\cite{DM}   & 2.78$\pm$0.14 & 37.3h                  & \textbf{2.92$\pm$0.15} & \textbf{29.6h} & \multirow{2}{*}{0.90$\pm$0.23} \\
        &                    & MTT~\cite{MTT} &  -            & 460.8h (\textit{est.}) & \textbf{2.77$\pm$0.21} & \textbf{76.8h} &  \\
        \cmidrule(l){2-8}                       
        &      10            &  DM~\cite{DM}  &  9.48$\pm$0.15 & 43.5h  &  \textbf{9.70$\pm$0.12}  &  \textbf{32.2h}  & 3.33$\pm$0.43     \\
        \bottomrule
      \end{tabular}
    } 
    \label{tab:large-data}
\end{table*}

\subsubsection{Higher-order Interaction of Data Utility} 
\label{subsec:high-order-utility}

Both our proposed preemptive and adaptive pruning is based on per-sample criteria, \ie we assume each sample independently contributes to the distillation process.
However, high-order information and interactions between samples may exist, and they have more complex causal mechanisms to the data synthesizing than the individual ITE in our method.
In some extreme scenarios, these interactions are not negligible.
For
\begin{wrapfigure}{r}{0.4\textwidth}
    \centering
    \includegraphics[width=0.95\linewidth]{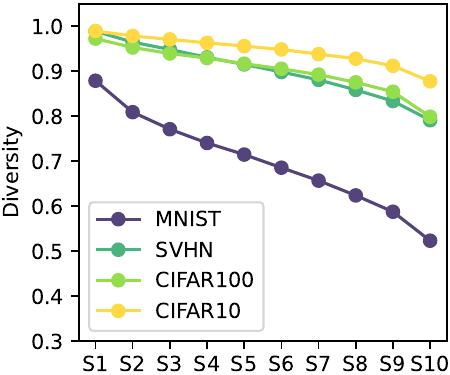} 
    \caption{The diversity of each stratified subgroup. MNIST suffers from diversity vanishing.}
    \label{fig:compare-diversity}
\end{wrapfigure}
example, \textit{data diversity} is a higher-order data utility indicator since it is a property of the population rather than an individual. 
We observe the data selection paradigm always performs poorly on low diversity datasets like MNIST, which is due to the \textit{diversity vanishing} in the small loss samples.
We conduct similar stratified experiments as \cref{sec:early-loss-ablation} and use the quotient of intraclass variance and interclass variance as the diversity metric (a large value indicates larger diversity).  
As shown in \cref{fig:compare-diversity}, only MNIST severely drops the diversity for the subgroups with a small loss.
It can be challenging to incorporate sample interactions like diversity into consideration without sacrificing efficiency (\eg use annealing or Monte-Carlo algorithms), so we leave it to future work. In-depth discussion and modeling of high-order data utility involve complex systems which is beyond our scope.
It is worth noting that the impact of diversity vanishing can be negligible in most realistic scenarios (\eg, CIFAR), especially for large-scale datasets due to their large overall diversity.
We provide the stratified visualization in the supplementary material.

\section{Conclusion}

This paper introduces a novel bi-level data pruning approach for efficient dataset distillation, which leverages the inherent redundancy in large datasets. We propose preemptive pruning from the dynamics of distillation, which is applied before the distillation, accompanied by further adaptive pruning based on causal effects. 
Our method would consistently enhance the performance of distillation algorithms while reducing the computational burden. 
We believe that our findings will inspire future research to explore more sophisticated data utility models and optimization techniques, ultimately leading to more efficient and effective dataset distillation methods.

\section*{Acknowledgments}
This work is supported in part by the
National Natural Science Foundation of China under Grants No.62306175.

%
%
\bibliographystyle{splncs04}
\bibliography{main}

\clearpage

\vbox{%
    \hsize\textwidth
    \linewidth\hsize
    \vskip 0.1in
    \centering
    {\LARGE\bf Supplementary Materials \par}
    \vskip 0.3in
}

We provide the following details and analyses in the supplementary:

Sec.~\ref{sec:loss-proof}: Relation between Loss and Its Derivative.

Sec.~\ref{sec:ite-error}: Error of ITE Estimation.

Sec.~\ref{sec:compare-coreset}: Comparison to Coreset Selection Methods.

Sec.~\ref{sec:overfit}: Overfitting Analysis.

Sec.~\ref{sec:cross-arch}: Model Generalization.

Sec.~\ref{sec:impl-details}: Implementation Details.

Sec.~\ref{sec:visualize}: More Visualizations.

Sec.~\ref{sec:license}: Licenses.

\section{Relation between Loss and Its Derivative}
\label{sec:loss-proof}

In BiLP, we use $\ell(u_r, y_r)$ as preemptive selection criterion since it is monotonous concerning $\|\frac{\partial \ell(u_r, y_r)}{\partial u_r}\|$ for common loss functions like Mean Squared Error (MSE) or cross-entropy, or say $\ell(u_r, y_r)$ and $\|\frac{\partial \ell(u_r, y_r)}{\partial u_r}\|$ are positive correlated.
Let's consider the two common loss functions mentioned:

\begin{enumerate}
\item \textbf{Mean Squared Error (MSE)}:
    The MSE loss function is defined as $\ell(u, y) = \frac{1}{2}(u - y)^2$, where $u$ is the predicted value and $y$ is the actual value. The gradient of the MSE with respect to $u$ is $\frac{\partial \ell}{\partial u} = u - y$. The gradient norm is then $\|\frac{\partial \ell}{\partial u}\| = |u - y|$.
    Thus, larger $\|\frac{\partial \ell}{\partial u}\|$ leads to larger$\ell(u, y)$.
\item \textbf{Cross-Entropy}:
    The cross-entropy loss function for binary classification is defined as $\ell(\boldsymbol{u}, y) = - \log(u_y)$, where $\boldsymbol{u}$ is the predicted logits of the $N$ class and each component $u_i>0$, and $y$ is the actual class. 
    The gradient of the cross-entropy with respect to each component $u_i$ is $\frac{\partial \ell}{\partial u_i} = \begin{cases}
        -\frac1{u_y}, & \text{if}~~i=y \\
        0, &otherwise
    \end{cases}$. The gradient norm is $\|\frac{\partial \ell}{\partial u}\| = \frac{1}{u_y\sqrt{N}}$.
    So larger $\|\frac{\partial \ell}{\partial u}\|$ indicating a small $u_y$ and therefore the $\ell(\boldsymbol{u}, y)$ is large.
\end{enumerate}

Overall, for these two loss functions, $\|\frac{\partial \ell}{\partial u_r}\|$ and $\ell(u_r, y_r)$ are positively correlated.

\section{Error of ITE Estimation}
\label{sec:ite-error}

\begin{figure*}[t]
    \centering
    \includegraphics[width=0.99\textwidth]{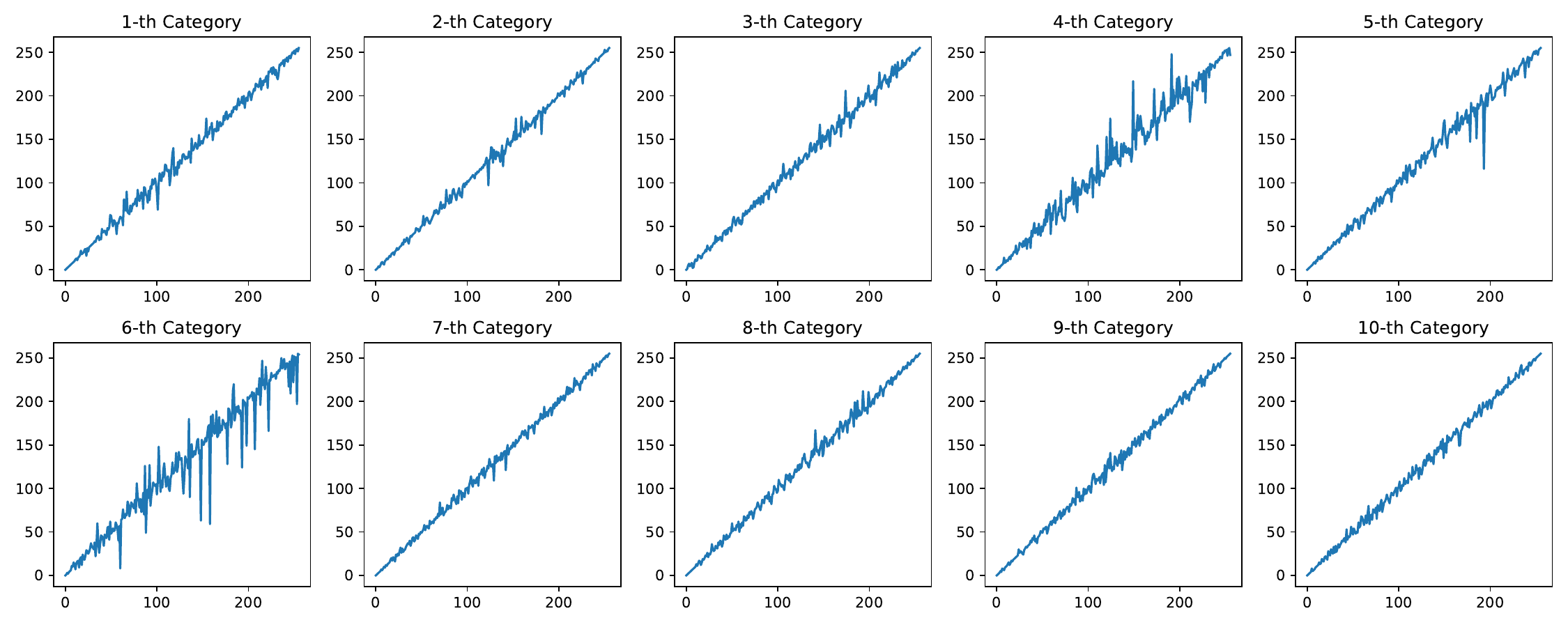}
    \caption{Visualization of the original and the Taylor-approximated ITE rankings for CIFAR10~\cite{cifar} with IPC=10. An ideal approximation would reveal a 45-degree diagonal line.}
    \label{suppfig:ite-approx}
\end{figure*}

In BiLP, we have implemented the Taylor approximation technique to enhance the efficiency of ITE value computations. To quantitatively assess the error of the Taylor approximation, we conducted an experiment where both the original and the Taylor-approximated ITE values were calculated within a mini-batch consisting of $N$ samples. The samples were then sorted based on their ITE values, resulting in rankings represented by $r_i$ and $r'_i$ for each $i^{\text{th}}$ sample within the range of $[1, N]$.
To evaluate the discrepancy between the two sets of rankings, we utilized the average relative error metric, defined as $\frac{1}{N(N-1)}\sum_{i=1}^N |r_i - r'_i|$, with a range of $[0, 1)$. Experimenting on the CIFAR10 dataset with an IPC of 10, our analysis revealed an average error rate of 2.7\%, which is considered negligible for pruning.

\begin{table*}[t]
    \centering
    \captionof{table}{Comparison to coreset selection on CIFAR10, IPC=1 and DC algorithm.}
    \label{tab:coreset-1}
    \resizebox{0.9\textwidth}{!}{
      \begin{tabular}{l|cccccc|c}
        \toprule
        \multirow{2}{*}{\makecell{Selection\\Criterion} } & \multicolumn{7}{c}{Pruning Ratio} \\
        \cmidrule(l){2-8}
         &   99\% & 97\% & 95\% & 90\% & 80\% & 70\% & 0\% (Full dataset)\\
        \midrule
        Random & 25.6$\pm$0.6 & 27.6$\pm$0.8 & 27.6$\pm$0.6 & 28.2$\pm$0.3 & 28.5$\pm$0.4 & 28.7$\pm$0.3 & \multirow{5}{*}{28.3$\pm$0.5} \\
        \cmidrule(l){1-7}
        Loss (remove large) & \textbf{29.7$\pm$0.1} & \textbf{29.7$\pm$0.0} & \textbf{30.0$\pm$0.1} & \textbf{30.0$\pm$0.2} & \textbf{30.2$\pm$0.1} & \textbf{30.2$\pm$0.2} & \\
        \cmidrule(l){1-7}
        Coreset: CRAIG~\cite{CoreCRAIG} & 25.4$\pm$0.6 & 27.8$\pm$0.2 & 28.8$\pm$0.5 & 28.6$\pm$0.4 & 29.0$\pm$0.1 & 29.0$\pm$0.2 \\
        Coreset: GradMatch~\cite{CoreGradMatch} & 26.5$\pm$0.5 & 28.0$\pm$0.3 & 28.7$\pm$0.5 & 28.4$\pm$0.3 & 28.9$\pm$0.2 & 29.4$\pm$0.3 & \\
        Coreset: GLISTER~\cite{CoreGLISTER} & 23.8$\pm$0.5 & 26.9$\pm$0.2 & 23.5$\pm$0.1 & 21.3$\pm$0.6 & 24.0$\pm$0.9 & 24.9$\pm$0.8 & \\
        \bottomrule
      \end{tabular}
    }
\end{table*}

Furthermore, to provide a more intuitive understanding of our approximation's performance, we visualize the rankings in \cref{suppfig:ite-approx}. The x-axis corresponds to the original ITE rankings, while the y-axis depicts the rankings derived from the Taylor approximation, with each of the 10 classes distinctly plotted. The closer the distribution of points to a 45-degree diagonal line, the higher the similarity between the two computation methods. The figure demonstrates that our Taylor approximation closely mirrors the original ITE rankings, thereby validating its efficacy.

\section{Comparison to Coreset Selection Methods}
\label{sec:compare-coreset}

\begin{table*}[t]
    \centering
    \captionof{table}{Comparison to coreset selection on CIFAR10, IPC=50 and DC algorithm.}
    \label{tab:coreset-50}
    \resizebox{0.7\textwidth}{!}{
      \begin{tabular}{l|cccc|c}
        \toprule
        \multirow{2}{*}{\makecell{Selection\\Criterion} } & \multicolumn{5}{c}{Pruning Ratio} \\
        \cmidrule(l){2-6}
         &   70\% & 60\% & 50\% & 30\% & 0\% (Full)\\
        \midrule
        Random 	            & 53.0$\pm$0.2 & 53.6$\pm$0.3 & 54.0$\pm$0.5 & 54.2$\pm$0.3 & \multirow{5}{*}{54.1$\pm$0.3} \\
        \cmidrule(l){1-5}
        Loss (remove large) & \textbf{54.1$\pm$0.2} & \textbf{54.9$\pm$0.3} & \textbf{55.3$\pm$0.3} & \textbf{56.0$\pm$0.2} & \\
        \cmidrule(l){1-5}
        Coreset: CRAIG~\cite{CoreCRAIG} 	& 48.8$\pm$0.4 & 48.8$\pm$0.2 & 49.0$\pm$0.4 & 49.1$\pm$0.4 & \\
        Coreset: GradMatch~\cite{CoreGradMatch} 	& 49.0$\pm$0.3 & 49.1$\pm$0.3 & 49.1$\pm$0.4 & 49.1$\pm$0.5 & \\
        Coreset: GLISTER~\cite{CoreGLISTER} 	& 41.9$\pm$0.4 & 43.9$\pm$0.5 & 44.0$\pm$0.5 & 47.0$\pm$0.5 & \\
        \bottomrule
      \end{tabular}
    }
\end{table*}

The existing coreset selection methods can also be exploited as sample selection criteria.
So we conduct a comparison with recent coreset selection methods on CIFAR10~\cite{cifar} and DC~\cite{DC} algorithm, including CRAIG~\cite{CoreCRAIG}, GradMatch~\cite{CoreGradMatch} and GLISTER~\cite{CoreGLISTER}. 
We adopt these methods as a preemptive pruning criterion.
We use the algorithms implemented by the CORDS package. The results are shown in Tab.~\ref{tab:coreset-1} and \ref{tab:coreset-50}. CRAIG and GradMatch coreset selection can achieve a 97\% maximum pruning ratio when IPC=1, though still worse than the loss criterion. The GLISTER algorithm does not perform well on dataset distillation and is worse than random selection. Thus, on the data pruning for dataset distillation, our loss indicator can surpass some sophisticated selection algorithms.

\begin{table}[t]
\begin{minipage}{0.38\textwidth}
    \centering
    \caption{Maximum pruning ratio on more distillation algorithms.}
    \label{tab:rand-drop-more-method}
    \resizebox{0.96\linewidth}{!}{
      \begin{tabular}{l|l|ccc}
        \toprule
        Dataset & IPC  & CAFE~\cite{CAFE} & LinBa~\cite{LinBa} & IDC~\cite{IDC}\\
        \midrule[1.3pt]
        \multirow{2}{*}{CIFAR10~\cite{cifar}}& 1  & 85\% & 30\% & 50\% \\
                                & 10 & 89\% & 70\% & 90\% \\
        \midrule
        \multirow{2}{*}{SVHN~\cite{svhn}}   & 1  & 70\% & 50\% & 40\% \\
                                & 10 & 40\% & 40\% & 70\% \\
        \midrule
        \multirow{2}{*}{MNIST~\cite{mnist}}  & 1  & 90\% & 70\% &  99.5\% \\
                                & 10 &  1\% & 60\% &  60\% \\
        \bottomrule
      \end{tabular}
    } 
\end{minipage}
\hfill
\begin{minipage}{0.28\textwidth}
    \centering
    \caption{Maximum pruning ratio of various initializations on CIFAR10~\cite{cifar}.}
    \label{tab:rand-drop-init}
    \resizebox{0.96\textwidth}{!}{
      \begin{tabular}{l|l|ccc}
        \toprule
        IPC  & Init &DC~\cite{DC} & DM~\cite{DM}\\
        \midrule[1.3pt]
        \multirow{3}{*}{1}  & Noise   & 90\% & 90\% \\
                            & Real    & 90\% & 85\% \\
                            & Herd    & 90\% & 85\% \\
        \midrule
        \multirow{3}{*}{10} & Noise   & 70\% & 70\% \\
                            & Real    & 70\% & 60\% \\
                            & Herd    & 70\% & 70\% \\
        \bottomrule
      \end{tabular}
    } 
\end{minipage}
\hfill
\begin{minipage}{0.3\textwidth}
    \centering
    \caption{Maximum pruning ratio of various networks on CIFAR10~\cite{cifar}.}
    \label{tab:rand-drop-network}
    \resizebox{\textwidth}{!}{
      \begin{tabular}{l|l|ccc}
        \toprule
        IPC  & Net &  DC~\cite{DC} & DM~\cite{DM}  \\
        \midrule[1.3pt]
        \multirow{5}{*}{1}  & Conv~\cite{DD}   & 90\% & 85\% \\
                            & MLP    &  97\% &  95\% \\
                            & ResNet~\cite{resnet} &  95\% & 85\% \\
                            & VGG~\cite{vgg}    & 90\% &  95\% \\
                            & AlexNet~\cite{alexnet}&  95\% &  95\% \\
        \midrule
        \multirow{2}{*}{10} & Conv~\cite{DD}   & 70\% & 60\% \\
                            & MLP    & 60\% & 60\% \\
        \bottomrule
      \end{tabular}
    } 
\end{minipage}
\end{table}

\subsection{Data Redundancy on Various Architectures and Initialization}
In the main paper, we examine the real data redundancy in dataset distillation by randomly removing some real samples before the training. 
Here we present more results in \cref{tab:rand-drop-more-method,tab:rand-drop-init,tab:rand-drop-network} with more algorithms and various initialization and network architectures, which exhibit large pruning ratios and indicate significant data redundancy.

\section{Overfitting Analysis}
\label{sec:overfit}

\begin{figure}[t]
    \centering
    \subfloat{
        \includegraphics[width=0.4\linewidth]{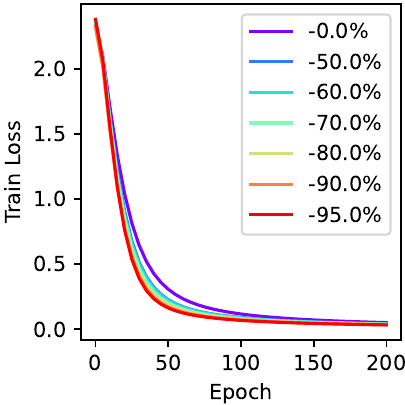}
    }
    \subfloat{
        \includegraphics[width=0.4\linewidth]{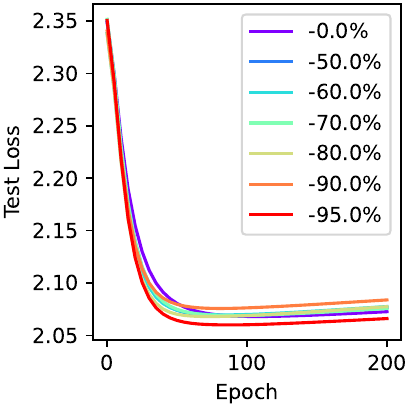}
    }
    \caption{Train and test loss curves for different pruning rates.}
    \label{supfig:overfit}
\end{figure}

We compare the train and test loss curve on distilled data with different pruning rates in Figure \ref{supfig:overfit} on CIFAR10 and IPC=1 (5 repeats). The test loss curves do not drastically increase after the convergence of training, and all the loss curves are similar, showing that a large pruning rate does not enhance overfitting problems.
This is consistent with the explanation in Sec.~4.1: due to the limited capacity of the synthetic dataset, removing some unimportant or outlier data samples does not harm the distillation process.

\section{Model Generalization}
\label{sec:cross-arch}

\subsection{Cross-Architecture Generalization of Preemptive Pruning}

We conducted a cross-architecture evaluation to verify whether the data pruning harms the generalization ability. 
We first follow DC~\cite{DC} to experiment on MNIST and IPC=1. We remove the training samples with the largest loss values and we compare the training subsets with 100\% (full data, original setting in DC paper), 10\%, 5\%, and 3\%. The results are shown in Tab.~\ref{tab:cross-arch-mnist} in the rebuttal PDF file, showing that pruning the training dataset does not damage the generalization ability of the distilled data. On the contrary, in most slots (28/36), data pruning can even enhance the generalization ability.

\begin{table*}[t]
    \centering
    \caption{Cross-architecture generalization on CIFAR10~\cite{cifar} and IPC=1 with DC~\cite{DC} algorithm. Best results are marked in \BGreen{bold green}. The generalization ability of pruned data is superior or comparable to the originals.}
    \label{tab:cross-arch-mnist}
    \resizebox{0.9\textwidth}{!}{
      \begin{tabular}{l|c|cccccc}
        \toprule
        \multirow{2}{*}{\makecell{Evaluate\\Network}} &  \multirow{2}{*}{\makecell{Pruning\\ Ratio}} & \multicolumn{6}{c}{Distill Network} \\
         &   & MLP & ConvNet & LeNet & AlexNet & VGG & ResNet\\
        \midrule[1.3pt]
        \multirow{4}{*}{MLP}
    &  0\% (Full) & 24.3\STD{1.8}  & 26.5\STD{0.2}  & 24.1\STD{0.4}  & 26.5\STD{0.1}  & 25.1\STD{0.3}  & 24.9\STD{0.4}  \\
           & 60\% & 23.0\STD{1.4}  & 26.6\STD{0.3}  & 24.2\STD{0.5}  & 26.6\STD{0.1}  & 25.1\STD{0.1}  & \BGreen{25.1\STD{0.1}}  \\
           & 80\% & 23.7\STD{0.9}  & 26.5\STD{0.4}  & 24.3\STD{0.7}  & 26.6\STD{0.1}  & 25.2\STD{0.1}  & 24.6\STD{0.3}  \\
           & 95\% & \BGreen{25.1\STD{1.4}}  & \BGreen{26.8\STD{0.2}}  & \BGreen{24.3\STD{0.2}}  & \BGreen{26.6\STD{0.2}}  & \BGreen{25.3\STD{0.3}}  & 24.8\STD{0.4}  \\
        \midrule
        \multirow{4}{*}{ConvNet~\cite{DD}}
    &  0\% (Full) & 22.7\STD{1.1}  & 28.8\STD{0.4}  & 21.8\STD{0.2}  & 27.0\STD{0.5}  & 25.8\STD{0.4}  & 26.2\STD{0.3}  \\
           & 60\% & 23.0\STD{1.7}  & 28.7\STD{0.1}  & 22.1\STD{0.3}  & 26.9\STD{0.2}  & 25.8\STD{0.3}  & 26.2\STD{0.2}  \\
           & 80\% & 23.4\STD{0.9}  & 28.8\STD{0.2}  & 22.2\STD{0.3}  & 26.8\STD{0.3}  & 25.6\STD{0.2}  & \BGreen{26.2\STD{0.2}}  \\
           & 95\% & \BGreen{23.7\STD{0.8}}  & \BGreen{29.1\STD{0.2}}  & \BGreen{22.3\STD{0.3}}  & \BGreen{26.9\STD{0.4}}  & \BGreen{25.8\STD{0.1}}  & 26.1\STD{0.3}  \\
        \midrule
        \multirow{4}{*}{LeNet~\cite{mnist}}
    &  0\% (Full) & 16.0\STD{2.6}  & 17.9\STD{0.8}  & 14.9\STD{1.1}  & 15.7\STD{1.5}  & 18.2\STD{1.3}  & 16.6\STD{1.1}  \\
           & 60\% & 16.8\STD{1.9}  & 17.6\STD{0.8}  & \BGreen{15.6\STD{0.2}}  & 16.1\STD{0.7}  & 18.8\STD{0.4}  & 16.4\STD{1.1}  \\
           & 80\% & \BGreen{18.9\STD{1.5}}  & 17.3\STD{0.6}  & 14.0\STD{0.6}  & \BGreen{16.2\STD{1.1}}  & \BGreen{19.3\STD{0.6}}  & 15.8\STD{0.9}  \\
           & 95\% & 17.9\STD{1.0}  & \BGreen{17.9\STD{1.2}}  & 15.1\STD{0.7}  & 15.6\STD{0.6}  & 19.1\STD{0.8}  & \BGreen{16.8\STD{1.2}}  \\
        \midrule
        \multirow{4}{*}{AlexNet~\cite{alexnet}}
    &  0\% (Full) & 20.8\STD{0.5}  & 22.9\STD{0.3}  & 22.7\STD{0.6}  & 21.1\STD{0.2}  & 22.7\STD{0.2}  & 22.3\STD{0.2}  \\
           & 60\% & 19.4\STD{1.2}  & 22.7\STD{0.4}  & 23.0\STD{0.4}  & 21.1\STD{0.4}  & 22.7\STD{0.1}  & 22.3\STD{0.2}  \\
           & 80\% & 21.2\STD{0.6}  & \BGreen{23.0\STD{0.3}}  & 23.1\STD{0.7}  & 20.6\STD{0.1}  & 22.6\STD{0.1}  & 22.2\STD{0.3}  \\
           & 95\% & \BGreen{21.3\STD{0.4}}  & 22.9\STD{0.2}  & \BGreen{23.1\STD{0.4}}  & \BGreen{21.2\STD{0.5}}  & \BGreen{22.8\STD{0.2}}  & \BGreen{22.3\STD{0.2}}  \\
        \midrule
        \multirow{4}{*}{VGG~\cite{vgg}}
    &  0\% (Full) & 20.3\STD{3.1}  & 15.7\STD{0.6}  & 20.4\STD{0.8}  & 23.6\STD{0.7}  & 18.0\STD{0.8}  & \BGreen{18.5\STD{0.4}}  \\
           & 60\% & 20.2\STD{1.6}  & \BGreen{15.8\STD{0.4}}  & 20.3\STD{0.5}  & \BGreen{24.0\STD{0.6}}  & \BGreen{18.6\STD{0.4}}  & 18.1\STD{0.3}  \\
           & 80\% & 19.9\STD{2.2}  & 15.3\STD{0.8}  & \BGreen{20.6\STD{0.8}}  & 23.6\STD{0.6}  & 18.1\STD{0.7}  & 18.2\STD{0.7}  \\
           & 95\% & \BGreen{20.4\STD{1.7}}  & 15.3\STD{0.7}  & 20.4\STD{0.8}  & 23.0\STD{0.8}  & 18.0\STD{0.9}  & 18.2\STD{0.7}  \\
        \midrule
        \multirow{4}{*}{ResNet~\cite{resnet}}
    &  0\% (Full) & 18.3\STD{1.6}  & 16.7\STD{0.9}  & 15.1\STD{0.8}  & 17.5\STD{0.5}  & 16.0\STD{0.7}  & 18.1\STD{1.1}  \\
           & 60\% & \BGreen{18.9\STD{1.8}}  & \BGreen{17.2\STD{1.0}}  & \BGreen{15.2\STD{0.5}}  & \BGreen{17.8\STD{0.6}}  &\BGreen{ 15.6\STD{1.0}}  & 18.5\STD{0.6}  \\
           & 80\% & 17.0\STD{1.1}  & 16.3\STD{0.7}  & 13.9\STD{1.0}  & 17.6\STD{1.3}  & 15.2\STD{0.6}  & \BGreen{19.2\STD{0.8}}  \\
           & 95\% & 18.8\STD{1.4}  & 16.7\STD{0.6}  & 14.3\STD{0.7}  & 17.0\STD{1.4}  & 15.0\STD{0.4}  & 18.3\STD{0.7}  \\
        \bottomrule
      \end{tabular}
    }
\end{table*}

\begin{table*}[t]
    \centering
    \caption{Cross-architecture generalization on CIFAR10~\cite{cifar} and IPC=50 with DC~\cite{DC} algorithm.}
    \label{tab:cross-arch-ipc50dc}
    \resizebox{0.9\textwidth}{!}{
      \begin{tabular}{l|ccccccc}
        \toprule
        \multirow{2}{*}{Sample Ratio} & \multicolumn{6}{c}{Evaluate Network} \\
                        & MLP & ConvNet & LeNet & AlexNet & VGG & ResNet\\
        \midrule[1.3pt]
        
        0\% (Full) & 28.01$\pm$0.40 & 54.02$\pm$0.51 & 28.12$\pm$2.16 & 29.48$\pm$0.58 & 39.44$\pm$0.67 & 22.72$\pm$1.08 \\
        30\%  & 29.48$\pm$0.28 & \textbf{55.96$\pm$0.40} & 30.83$\pm$1.51 & 29.54$\pm$2.60 & 41.99$\pm$0.47 & 24.35$\pm$0.42 \\
        50\%  & \textbf{30.40$\pm$0.28} & 55.25$\pm$0.32 & 31.15$\pm$1.25 & 30.53$\pm$2.21 & 43.09$\pm$0.41 & \textbf{25.81$\pm$1.04} \\
        70\%  & 30.15$\pm$0.21 & 54.77$\pm$0.47 & \textbf{31.44$\pm$0.72} & \textbf{33.45$\pm$1.18} & \textbf{43.35$\pm$0.60} & 25.48$\pm$0.53 \\
        \bottomrule
      \end{tabular}
    }
\end{table*}

We also conduct experiments on larger IPCs with DC~\cite{DC} and MTT~\cite{MTT}. The results are shown in Tab.~\ref{tab:cross-arch-ipc50dc} and ~\ref{tab:cross-arch-ipc10mtt}. On larger IPCs, pruning the training dataset still does not damage the generalization ability of the distilled data.

\begin{table*}[t]
    \centering
    \caption{Cross-architecture generalization on CIFAR10~\cite{cifar} and IPC=10 with MTT~\cite{MTT} algorithm.}
    \label{tab:cross-arch-ipc10mtt}
    \resizebox{0.6\textwidth}{!}{
      \begin{tabular}{l|ccccc}
        \toprule
        \multirow{2}{*}{Sample Ratio} & \multicolumn{4}{c}{Evaluate Network} \\
                        & ConvNet & AlexNet & VGG & ResNet\\
        \midrule[1.3pt]
        0\% (Full)  & 64.3$\pm$0.7 & 34.2$\pm$2.6 & 50.3$\pm$0.8 & 46.4$\pm$0.6 \\
        10\%   & \textbf{64.6$\pm$0.4} & \textbf{34.3$\pm$2.4} & \textbf{51.1$\pm$1.1} & \textbf{48.6$\pm$0.4} \\
        \bottomrule
      \end{tabular}
    }
\end{table*}

\begin{table*}[t]
\centering
\caption{Cross-architecture evaluation of synthetic data trained on ConvNet-D3 with BiLP, on CIFAR10~\cite{cifar}.}
\label{tab:cross-arch-bilp}
\resizebox{0.6\textwidth}{!}{
    \begin{tabular}{l|l|ccc}
        \toprule
        \multirow{2}{*}{Architecture} & \multirow{2}{*}{Method} & \multicolumn{3}{c}{IPC} \\
         & & 1 & 10 & 50 \\
        \midrule
        \multirow{2}{*}{ConvNet-D3~\cite{DD}}  
            & IDC      & 50.6\STD{0.4} & 67.5\STD{0.5} & 74.5\STD{0.1} \\
            & BiLP+IDC & \textbf{51.5\STD{0.3}} & \textbf{69.4\STD{0.5}} & \textbf{75.4\STD{0.2}} \\
        \midrule
        \multirow{2}{*}{ResNet~\cite{resnet}}      
            & IDC      & 42.8\STD{1.2} & 64.8\STD{0.9} & 71.9\STD{1.3} \\
            & BiLP+IDC & \textbf{43.3\STD{0.6}} & \textbf{65.5\STD{1.0}} & \textbf{72.8\STD{0.4}} \\
        \midrule
        \multirow{2}{*}{EfficientNet~\cite{efficientnet}}
            & IDC      & 38.5\STD{1.0} & 42.2\STD{0.8} & 71.4\STD{1.5} \\
            & BiLP+IDC & \textbf{38.5\STD{0.6}} &\textbf{ 43.7\STD{2.5}} & \textbf{71.9\STD{1.0}} \\
        \midrule
        \multirow{2}{*}{DenseNet~\cite{densenet}}
            & IDC      & 37.9\STD{0.6} & \textbf{64.8\STD{0.8}} & 70.4\STD{0.5} \\
            & BiLP+IDC & \textbf{38.3\STD{0.5}} & 64.6\STD{0.4} & \textbf{70.7\STD{0.7}} \\
        \bottomrule
      \end{tabular}
    }
\end{table*}

\subsection{Cross-Architecture Generalization of Full BiLP}

In this study, we present a cross-architecture evaluation that demonstrates the efficacy of synthetic data trained on the ConvNet-D3 architecture while being assessed across different neural network architectures including ResNet~\cite{resnet}, EfficientNet~\cite{efficientnet}, and DenseNet~\cite{densenet}. The experiments are conducted using the CIFAR10 dataset at three IPC levels 1, 10, and 50.
As shown in \cref{tab:cross-arch-bilp}, the synthetic data generalize well when applied to other network architectures, \eg at IPC=50, all three model types achieve performance that is on par with those obtained from the in-situ evaluation using ConvNet-D3.
This indicates that the synthetic data is capable of maintaining its predictive accuracy across diverse architectures.
Furthermore, our comparative analysis reveals that the BiLP consistently outperforms IDC in the majority of the conducted experiments.

\subsection{Generalization to Larger Data}

We add experiments on larger dataset (\eg ImageNet) and larger IPC (we adopt DATM~\cite{DATM} algorithm). We re-implement DATM with TESLA for efficiency and compare BiLP (preemptive pruning 10\% data) to the reproduced DATM results. 
As shown in Table \ref{suptab:datm}, BiLP could enhance DATM on small IPCs and is comparable at large IPCs, and could also achieve lossless performance at IPC=1000. We also conduct experiment of SRe2L~\cite{sre2l} in Table \ref{suptab:sre2l}. BiLP could enhance the performance with less data, \textit{e.g.}, BiLP with only 50\% data could significantly enhance SRe2L.

\begin{table}[t]
\begin{minipage}{0.49\textwidth}
    \centering
    \caption{DATM~\cite{DATM} and BiLP performance on CIFAR10. Full=84.8\%.}
    \resizebox{\linewidth}{!}{
      \begin{tabular}{l|ccccc}
        \toprule
        IPC        &    1  & 10   &  50  &   500 & 1000 \\
        \midrule[1.3pt] 
        DATM       &  47.0 & 65.7 & 72.9 & \textbf{81.3}  &  \textbf{84.8}  \\
        DATM+BiLP  &  {\bf47.4} & {\bf66.1} & {\bf74.0} & \textbf{81.3}  & 84.6 \\
        \bottomrule
      \end{tabular}
    } 
    \label{suptab:datm}
\end{minipage}
\hfill
\begin{minipage}{0.48\textwidth}
    \centering
    \caption{SRe2L~\cite{sre2l} and BiLP performance on large dataset ImageNet~\cite{imagenet}.}
    \resizebox{0.9\linewidth}{!}{
      \begin{tabular}{l|cc}
        \toprule
        IPC          &    1  & 10   \\
        \midrule[1.3pt] 
        SRe2L        &  1.2  & 21.3  \\ 
        SRe2L+BiLP (50\% data)   &  \textbf{1.6}   & \textbf{27.0}  \\
        \bottomrule
      \end{tabular}
    }
    \label{suptab:sre2l}
\end{minipage}
\end{table}

\section{Implementation Details}
\label{sec:impl-details}

\subsection{Datasets and Metric}

Our experiments are conducted on the following datasets and we report the top-1 accuracy as the metric, most of which are widely adopted in dataset distillation.

\begin{itemize}
    \item CIFAR10~\cite{cifar}: image dataset of common objects with 10 classes and 50,000 image samples. The images are 32x32 with 3 channels.
    \item CIFAR100~\cite{cifar}: image dataset of common objects with 100 classes and 50,000 samples. The images are 32x32 with 3 channels.
    \item SVHN~\cite{svhn}: street digit dataset with 10 classes and 73,257 samples. The images are 32x32 with 3 channels.
    \item TinyImageNet~\cite{tinyimgnet}: a subset of ImageNet with 200 classes and 100,000 images. The images are 64x64 with 3 channels.
    \item ImageNet~\cite{imagenet}: image datasets of common objects with 1000 classes and 1,281,167 samples. We resize the images to 64x64 with 3 channels following the previous setting~\cite{FRePo}.
    \item Kinetics-400~\cite{kinetics}: human action video dataset with 400 classes and 215,617 video samples. The videos are resampled to 8 frames per clip and resized to 64x64.
\end{itemize}

\subsection{Network Architectures}

Following the previous work, in most of the experiments, we adopt ConvNetD3 as the network to probe the data. This network consists of 3 convolutional layers with a 3x3 filter, each of which has 128 channels and is followed by a ReLU non-linearity and an InstaceNorm layer. The average pooling layer aggregates the feature map to a 128d vector and produces the logit with a linear layer.

We also adopt other architectures, including MLP (three linear layers with hidden layer size 128), AlexNet~\cite{alexnet}, ResNet18~\cite{resnet} (ResNet18+BatchNorm with average pooling for DC algorithm), and VGG11~\cite{vgg} (we use VGG11+BatchNorm for DC algorithm).

\subsection{Experiments of Random or Loss Selection}

In Tab.~1-3 in the main paper, we extensively study the critical sample ratio by random or loss value.
We mainly follow the default experiment settings given by each algorithm.
The experiments are conducted on RTX 4090 GPU.
We list the experiment details: 

\begin{enumerate}
    \item For DC~\cite{DC} and DSA~\cite{DSA}, on all datasets, we run the distillation for 1000 iterations with SGD optimizer and momentum 0.5. The number of inner loop and outer loop are (1, 1) for IPC=1, (10, 50) for IPC=10, (50, 10) for IPC=50. The learning rate of synthetic image and network are 0.1 and 0.01. The batch size for each class is 256 and when the sample ratio is low, we half the batch size until it is less than twice the largest class size. We use \texttt{color, crop, cutout, scale, rotate} DSA augmentation on all datasets and additional \texttt{flip} on the non-digit datasets. By default, \texttt{noise} initialization is used.
    \item For DM~\cite{DM}, we run the distillation for 10000 iterations on TinyImageNet and 20000 iterations for the others with SGD optimizer and momentum 0.5. The learning rate of synthetic image and network are 1.0 and 0.01. The batch size for each class is 256 and when the sample ratio is low, we half the batch size until it is less than twice the largest class size. The same Siamese augmentation strategy is used as in the DSA experiments. By default, \texttt{real} initialization is used (the initial images are drawn after dropping).
    \item For MTT~\cite{MTT}, we drop the same data samples for buffering and distillation. The expert trajectories are trained for 50 epochs for 100 repeats and we run the distillation for 10000 iterations. We appreciate and follow the \href{https://user-images.githubusercontent.com/18726777/184226412-7bd0d577-225b-487c-8c9c-23f6462ca7d0.png}{detailed hyper-parameters} provided by the authors.
    \item For CAFE~\cite{CAFE}, as default, we run the distillation for 2000 iterations. The initial learning rate is 0.1 and decays by 0.5 at 1,200, 1,600, and 1,800 iterations. The weight of the inner layer matching loss is 0.01 and an additional loss weight of 0.1 is put on the matching loss of the third and fourth layers. \texttt{Noise} initialization is used.
    \item For LinBa~\cite{LinBa}, we run distillation for 5000 iterations with SGD optimizer with momentum 0.9. The inner steps of BPTT are 150 and the number of bases is 16. The learning rate of synthetic image and network is 0.1 and 0.01.
    \item For IDC~\cite{IDC}, we use the ``reproduce'' setting of the opened source code, which automatically sets up the tuned hyper-parameters. We use multi-formation factor 2.
    \item For KIP~\cite{KIP}, we test on the finite-width model (KIP-NN) and use label learning. We use longer training steps for converged results.
    \item For FRePo~\cite{FRePo}, we use the official PyTorch implementation and the default parameters, except the learning rate of 0.001 and we run the distillation for 500,000 steps.
    \item For HaBa~\cite{haba}, we follow the official instructions and adopt the parameters from MTT. And for the exclusive parameters for HaBa, we use the values given in the code.
    \item For RFAD~\cite{RFAD}, we test on the finite-width model (ConvNet) and load the training hyperparameters for finite training results in the paper. The choice of label learning follows the remarks in the paper.
    \item For IDM~\cite{IDM}, we thank the authors and we directly adopt the official running commands.
\end{enumerate}

\textbf{The removal of data samples is class-wise}. Each experiment is repeated 5 times for mean $\mu$ and standard deviation $\sigma$ and we regard the experiment performance as comparable to the experiment on full data if its mean accuracy is within the $[\mu-\sigma, \mu+\sigma]$ of full data performance.

\subsection{Computation of Empirical Loss Criterion}
\label{subsec:impl-details-indicators}

The parameters and settings of the empirical loss criterion for the preemptive pruning are as follows:
We train the ConvNetD3 model on each dataset (ConvNetD3+GRU for Kinetics-400) for multiple trials for the loss indicator. We take the average loss curve of multiple trials. By default, we use a Gaussian filter with $\sigma=3$ to smooth the loss curve and take the loss value at the last epoch, which is approximately equivalent to the \textbf{weighted mean loss value of the last 8 epochs}. The training details are:

\begin{itemize}
    \item CIFAR10: 50 trials for 100 epochs with learning rate 3.0e-3 and batch size 512.
    \item CIFAR100: 50 trials for 250 epochs with learning rate 5.0e-3 and batch size 512.
    \item MNIST: 50 trials for 50 epochs with learning rate 3.0e-4 and batch size 512.
    \item SVHN: 50 trials for 100 epochs with learning rate 1.0e-3 and batch size 512.
    \item TinyImageNet: 50 trials for 100 epochs with learning rate 5.0e-3 and batch size 512.
    \item ImageNet: 30 trials for 20 epochs with learning rate 3.0e-3 and batch size 256 (early stop).
    \item Kinetics-400: 10 trials for 20 epochs with learning rate 1.0e-2 and batch size 128 (early stop).
\end{itemize}

Note that considering the conclusion in Sec.~5.5, we have adopted an early stop on large-scale datasets to reduce the training cost.

\subsection{Experiments of BiLP}

We use 3-layer ConvNet for CIFAR and SVHN datasets, and 4-layer ConvNet for TinyImageNet.
For BiLP, the momentum for running stats of ITE is set to 0.1 and we set $\beta=30\%$ by default. The other hyperparameters vary among different datasets and please refer to the supplementary.
We take the mean and standard deviation of the accuracy of 5 random trials.
Specifically, we show the hyper-parameters of different experiment settings in \cref{tab:param-bilp}.
We use the same parameters for multi-formation factor 2 or 3 for IDC.

\begin{table*}[t]
\centering
\caption{Hyper-parameters of BiLP in different experiments.}
\label{tab:param-bilp}
\resizebox{0.9\textwidth}{!}{
    \begin{tabular}{l|ccc|cc|ccc}
        \toprule
        Dataset & \multicolumn{3}{c|}{CIFAR10~\cite{cifar}} & \multicolumn{2}{c}{CIFAR100~\cite{cifar}} & \multicolumn{3}{c|}{SVHN~\cite{svhn}} \\
        IPC & 1 & 10 & 50  & 1 & 10 & 1 & 10 & 50   \\
        \midrule
        Preemptive pruning rate $\alpha$      & 0.1 & 0.3 & 0.1 & 0.6 & 0.4 & 0.2 & 0.3 & 0.3 \\
        Adaptive pruning rate $\beta$         & 0.3 & 0.3 & 0.3 & 0.3 & 0.3 & 0.2 & 0.2 & 0.2 \\
        Data update frequency (lazy selection) & 10  & 5   &  5  &  10 &  10 &  10 &  5  &  5  \\
        \bottomrule
      \end{tabular}
    }
\end{table*}


\subsection{Experiments on the Large-scale Datasets}

We apply our selection paradigm on larger-scale datasets in Sec.~5.2 in the main paper. The experiments are conducted on at most 4 RTX 3090 GPUs and the details are as follows:

\begin{itemize}
  \item ImageNet, DC: the training of DC exceeds the usual GPU capacity so in compromise we separate the 1000 classes into ten 100 class splits, which will slightly decrease the accuracy. The other hyper-parameters are the same as the previous experiments. For our paradigm, we prune 50\% samples and early stop at 800 iterations due to its faster convergence.
  \item ImageNet, DM: the DM algorithm is safe for class-separate training so we separate the classes into 4 splits at IPC=1, 8 splits at IPC=10, and 20 splits at IPC=50. We run the distillation for 5000 iterations with a learning rate of 5.0. For our paradigm, we prune 50\% samples and early stop at 2,000, 3,000, and 3,000 iterations for IPC=1/10/50 respectively.
  \item ImageNet, MTT: the expert trajectory is too expensive to compute so we only run MTT with our selection method. We prune 90\% samples which reduces 84\% of the trajectory training time. We train 60 trajectories for 50 epochs. MTT also requires large GPU memory due to the unrolling of back-propagation, so we use synthetic steps=5, expert epochs=2, and maximum start epoch=5. We run the distillation for 5000 iterations with an image learning rate of 30,000 and a step size learning rate of 1.0e-6.
  \item Kinetics-400, DM: on Kinetics, we run DM for 5000 iterations with a learning rate of 5.0 and batch size of 128. We separate the classes into 8 splits at IPC=1 and 20 splits at IPC=10. For our paradigm, we prune 50\% samples and early stop at 4000 iterations. We do not use DSA augmentation for Kinetics.
  \item Kinetics-400, MTT: we prune 90\% samples and train 40 trajectories for 50 epochs with batch size 128. We use synthetic steps=5, expert epochs=2, and maximum start epoch=5. We run the distillation for 5000 iterations with an image learning rate of 30,000, step size learning rate of 1.0e-6, real batch size 128, and synthetic batch size 64. We do not use DSA augmentation for Kinetics.
\end{itemize}

\section{More Visualizations}
\label{sec:visualize}

\begin{figure*}[ht]
    \centering
    \subfloat[0.43\textwidth][CIFAR100, loss indicator, small utility]{
        \centering
        \includegraphics[width=0.43\textwidth]{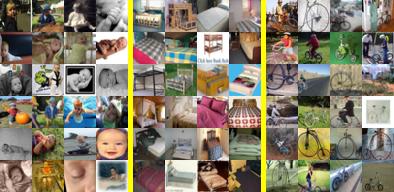}
    }
    \subfloat[0.43\textwidth][CIFAR100, loss indicator, large utility]{
        \centering
        \includegraphics[width=0.43\textwidth]{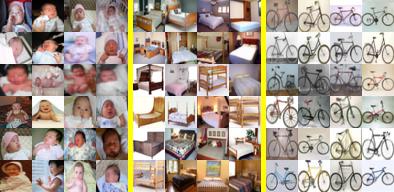}
    }

    \subfloat[0.43\textwidth][MNIST, loss indicator, small utility]{
        \centering
        \includegraphics[width=0.43\textwidth]{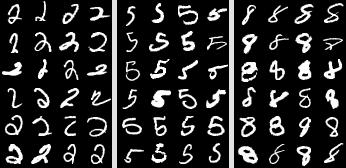}
    }
    \subfloat[0.43\textwidth][MNIST, loss indicator, large utility]{
        \centering
        \includegraphics[width=0.43\textwidth]{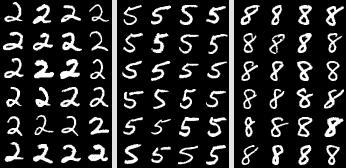}
    }

    \subfloat[0.43\textwidth][SVHN, loss indicator, small utility]{
        \centering
        \includegraphics[width=0.43\textwidth]{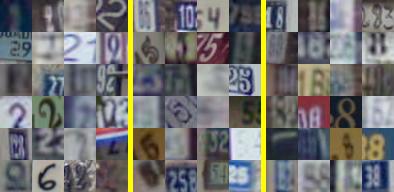}
    }
    \subfloat[0.43\textwidth][SVHN, loss indicator, large utility]{
        \centering
        \includegraphics[width=0.43\textwidth]{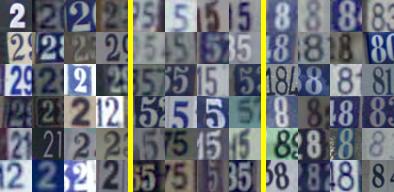}
    }

    \subfloat[0.43\textwidth][ImageNet, loss indicator, small utility]{
        \centering
        \includegraphics[width=0.43\textwidth]{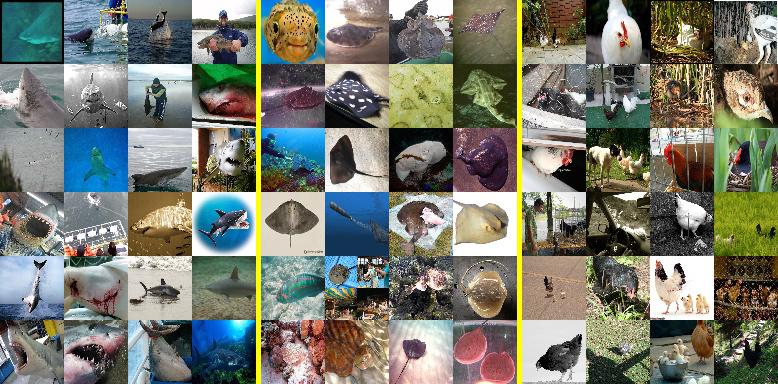}
    }
    \subfloat[0.43\textwidth][ImageNet, loss indicator, large utility]{
        \centering
        \includegraphics[width=0.43\textwidth]{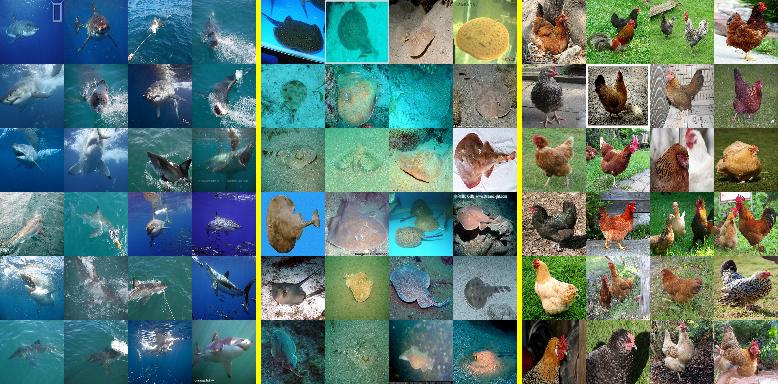}
    }

    \subfloat[0.43\textwidth][Kinetics-400, loss indicator, small utility]{
        \centering
        \includegraphics[width=0.43\textwidth]{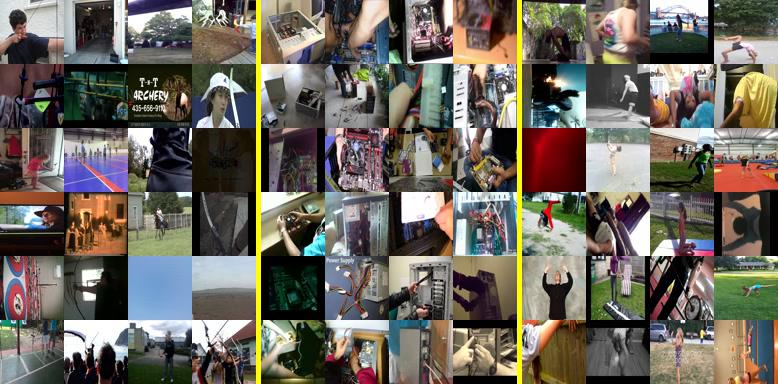}
    }
    \subfloat[0.43\textwidth][Kinetics-400, loss indicator, large utility]{
        \centering
        \includegraphics[width=0.43\textwidth]{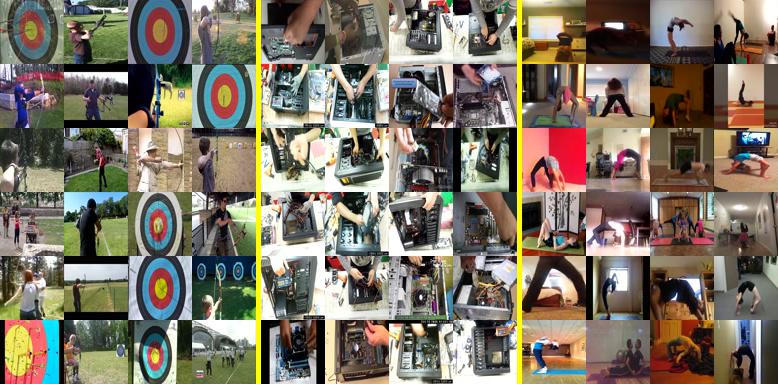}
    }
    \caption{Qualitative comparison of multiple datasets. We conduct stratified experiments with loss indicators and show samples in the layers with the smallest utility (left column) or largest utility (right column). We show three classes for each dataset.}
    \label{fig:example-various-datasets}
\end{figure*}

\begin{figure*}[t]
    \centering
    \subfloat[0.24\textwidth][Class ``1'', S1]{
        \centering
        \includegraphics[width=0.24\textwidth]{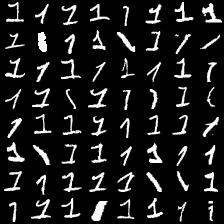}
    }
    \subfloat[0.24\textwidth][Class ``1'', S4]{
        \centering
        \includegraphics[width=0.24\textwidth]{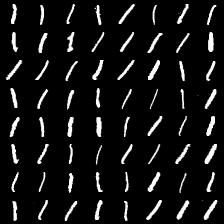}
    }
    \subfloat[0.24\textwidth][Class ``1'', S7]{
        \centering
        \includegraphics[width=0.24\textwidth]{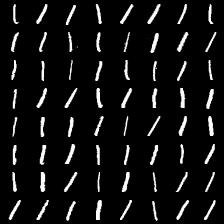}
    }
    \subfloat[0.24\textwidth][Class ``1'', S10]{
        \centering
        \includegraphics[width=0.24\textwidth]{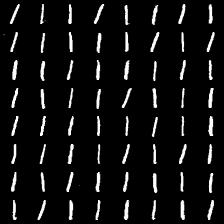}
    }

    \subfloat[0.24\textwidth][Class ``7'', S1]{
        \centering
        \includegraphics[width=0.24\textwidth]{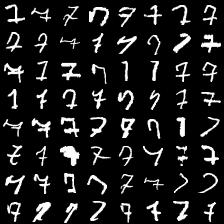}
    }
    \subfloat[0.24\textwidth][Class ``7'', S4]{
        \centering
        \includegraphics[width=0.24\textwidth]{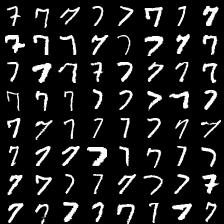}
    }
    \subfloat[0.24\textwidth][Class ``7'', S7]{
        \centering
        \includegraphics[width=0.24\textwidth]{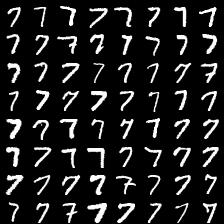}
    }
    \subfloat[0.24\textwidth][Class ``7'', S10]{
        \centering
        \includegraphics[width=0.24\textwidth]{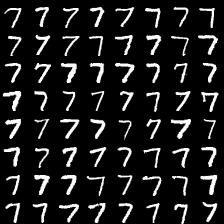}
    }
    \caption{More examples of different strata in the MNIST dataset. The data are stratified by classification loss. The samples in S1 have the lowest loss values and those in S10 have the largest loss. The diversity significantly drops when the sample loss decreases (\eg S7, S10).}
    \label{fig:example-diversity-mnist}
\end{figure*}

In this section, we present some data samples at different loss levels to qualitatively visualize the selection criterion.

We first stratify various datasets into 10 layers according to per-sample loss values and visualize some samples in the layer with the smallest or largest utility in Fig.~\ref{fig:example-various-datasets}, including the large-scale datasets (ImageNet and Kinetics-400).
As shown in the figure, the samples with small loss are noisy and usually hard and corner cases, \eg only part of the birds are shown, some dogs are acting in strange poses, or the images of ships are captured with unusual viewing angles. Meanwhile, the samples with large losses are easy cases that have ideal saliency, viewing angle, and clean background.

The digit datasets (MNIST, SVHN) show significantly more \textit{diversity vanishing} than the rest of the realistic datasets. Moreover, the diversity vanishing issue is mild for large-scale datasets such as ImageNet since the intra-class discrepancy is large such that any subset is diversified enough.

To extend our discussion on the data diversity (Sec.~5.4 in the main paper), we give some more examples to compare the diversity for data strata with different loss values in Fig.~\ref{fig:example-diversity-mnist} on MNIST. The groups with large loss values are mainly corner cases. Furthermore, as the loss value decreases (S7 or S10), the diversity significantly drops as shown in Fig.~\ref{fig:example-diversity-mnist} (c, d, g, h).

\section{Licenses}
\label{sec:license}

Here are the source and license of the assets involved in our work. We sincerely appreciate and thank the authors and creators. 

\noindent{\bf Datasets:}

\begin{itemize}
    \item CIFAR10, CIFAR100~\cite{cifar}: \href{https://www.cs.toronto.edu/~kriz/cifar.html}{URL}, unknown license.
    \item MNIST~\cite{mnist}: \href{http://yann.lecun.com/exdb/mnist/}{URL}, MIT License.
    \item SVHN~\cite{svhn}: \href{http://ufldl.stanford.edu/housenumbers/}{URL}, unknown license.
    \item Tiny-ImageNet~\cite{tinyimgnet}: \href{https://www.kaggle.com/competitions/tiny-imagenet/}{URL}, unknown license.
    \item ImageNet~\cite{imagenet}: \href{https://www.image-net.org}{URL}, custom license, research, non-commercial.
    \item Kinetics-400~\cite{kinetics}: \href{https://www.deepmind.com/open-source/kinetics}{URL}, Creative Commons Attribution 4.0 International License.
\end{itemize}

\noindent{\bf Code:}

\begin{itemize}
    \item DC~\cite{DC}, DSA~\cite{DSA}, DM~\cite{DM}: \href{https://github.com/VICO-UoE/DatasetCondensation}{URL}, MIT License.
    \item MTT~\cite{MTT}: \href{https://github.com/GeorgeCazenavette/mtt-distillation}{URL}, MIT License.
    \item CAFE~\cite{CAFE}: \href{https://github.com/kaiwang960112/CAFE}{URL}, no license.
    \item LinBa~\cite{LinBa}: \href{https://github.com/princetonvisualai/RememberThePast-DatasetDistillation}{URL}, no license.
    \item IDC~\cite{IDC}: \href{https://github.com/snu-mllab/Efficient-Dataset-Condensation/}{URL}, MIT License.
    \item KIP~\cite{KIP}: \href{https://colab.research.google.com/github/google-research/google-research/blob/master/kip/KIP.ipynb}{URL}, no license. 
    \item FRePo~\cite{FRePo}: \href{https://github.com/yongchao97/FRePo}{URL}, no license.
    \item HaBa~\cite{haba}: \href{https://github.com/Huage001/DatasetFactorization}{URL}, Apache-2.0 License.
    \item IDM~\cite{IDM}: \href{https://github.com/uitrbn/idm}{URL}, no license.
    \item RFAD~\cite{RFAD}: \href{https://github.com/yolky/RFAD}{URL}, no license.
    \item CORDS: \href{https://github.com/decile-team/cords}{URL}, MIT license.
\end{itemize}

\end{document}